%% file: sample-sigconf.tex
\documentclass[sigconf]{acmart}

\AtBeginDocument{%
  \providecommand\BibTeX{{%
    \normalfont B\kern-0.5em{\scshape i\kern-0.25em b}\kern-0.8em\TeX}}}

\setcopyright{acmcopyright}
\copyrightyear{2022}
\acmYear{2022}

\input{Match-commands}

\usepackage{bm}
\usepackage{wrapfig}
\usepackage{makecell}
\usepackage{csquotes}
\usepackage{subfigure}

\begin{document}

\title{Clutter Detection and Removal by Multi-Objective Analysis for Photographic Guidance}


\author{Xiaoran Wu}
\email{wuxr17@tsinghua.org.cn}
\affiliation{%
  \institution{Department of Computer Science and Technology, Tsinghua University}
  \city{Beijing}
  \country{China}
}

\renewcommand{\shortauthors}{Xiaoran Wu}
\settopmatter{printacmref=false}
\setcopyright{none}
\renewcommand\footnotetextcopyrightpermission[1]{}
\pagestyle{plain}
\begin{abstract}
Clutter in photos is a distraction preventing photographers from conveying the intended emotions or stories to the audience. Photography amateurs frequently include clutter in their photos due to unconscious negligence or the lack of experience in creating a decluttered, aesthetically appealing scene for shooting. We are thus motivated to develop a camera guidance system that provides solutions and guidance for clutter identification and removal. We estimate and visualize the contribution of objects to the overall aesthetics and content of a photo, based on which users can interactively identify clutter. Suggestions on getting rid of clutter, as well as a tool that removes cluttered objects computationally, are provided to guide users to deal with different kinds of clutter and improve their photographic work. Two technical novelties underpin interactions in our system: a clutter distinguishment algorithm with aesthetics evaluations for objects and an iterative image inpainting algorithm based on generative adversarial nets that reconstructs missing regions of removed objects for high-resolution images. User studies demonstrate that our system provides flexible interfaces and accurate algorithms that allow users to better identify distractions and take higher quality images within less time.
\end{abstract}




\maketitle

\input{1-Introduction}
\input{2-Related}
\input{3-Design}
\input{3-Method}
\input{5-Experiments}
\input{6-Conclusion}

\bibliographystyle{ACM-Reference-Format}
\bibliography{sample-base}


\end{document}

%% file: Match-commands.tex
\usepackage{amsmath,amsfonts,bm}

\def\eqref#1{equation~\ref{#1}}

\def\1{\bm{1}}

\DeclareMathAlphabet{\mathsfit}{\encodingdefault}{\sfdefault}{m}{sl}
\SetMathAlphabet{\mathsfit}{bold}{\encodingdefault}{\sfdefault}{bx}{n}



%% file: 1-Introduction.tex
\section{Introduction}

Photography can convey the emotions of photographers~\cite{kurtz2013happiness} to the audience by capturing a memorable moment, shooting a meaningful scene, or presenting an impressive portrait. In this regard, photographs serve a similar purpose to storytelling~\cite{block2020visual,cohn2013visual,williams2019attending}. However, stories can guide the feelings of an audience by giving larger context, while the observer of a photograph has to discover the context through visual clues~\cite{eisner2008graphic}. Clutter in a photo distracts the audience from capturing intended clues and may spoil the story told by the photo. Therefore, selecting a strong object and keeping visual clutter to a minimum are two critical factors for aesthetically appealing and emotion-invoking photos.

Amateur photographers frequently make the mistake of unexpectedly containing too much clutter in photos. There are many reasons for this mistake. Photographers may get accustomed to and unconsciously tune out visual clutter after staying in the environment for a while, or they may just ignore the clutter when focusing on other aspects of the photo, such as the major object or the configuration. Although it is typical of photographers to let clutter sneak into their photos, the audience is sensitive to every detail of the photo, and even a small cluttered object can form a considerable distraction~\cite{school_2013,blessing2002gina}.

The frequent appearance of visual clutter motivates camera interfaces~\cite{wu2021tumera,mitarai2013interactive} that remind the photographers of clutter in photos~\cite{jane2021dynamic}. However, a prompt reminder may not always satisfy users' needs. After being aware of the clutter, users typically try to remove clutter. However, removing clutter may not be easy. The gap between some cluttered objects and meaningful ones is close, and it requires experience or domain knowledge to identify this kind of clutter and form an aesthetically appealing composition. Moreover, even an experienced photographer may encounter clutter that is difficult to remove. Examples include numerous or immovable items in a scene, vehicles on a busy street, or endless tourists in a resort.

Closing the gap between users' needs of better dealing with clutter and the capability of current photography guidance systems, we develop a capture-time camera interface that provides photographers with solutions and guidance for clutter identification and removal. Specifically, we computationally estimate and visualize the contribution of objects to the overall aesthetic quality and content of the photo. Based on this quantification, we provide flexible interfaces for users to interactively identify clutter in the viewfinder. Furthermore, we provide suggestions and tools to remove different kinds of clutter. For outlier clutter, suggestions include zooming in, changing camera position, and changing orientation from portrait to landscape. For the clutter that cannot be easily removed, we provide an image inpainting tool that generates a preview image where the clutter is replaced with visually realistic and semantically reasonable background textures. Users can use this preview image as the final photographic work or as a reference or consideration when deciding whether it is worth the effort to remove the clutter in the real scene.

Two technical novelties underpin these interactions. First, we propose a novel deep neural network architecture for estimating the contribution of each object to the aesthetics and content. The key idea here is learning the overall aesthetic and content score of an image as a decomposition of sub-images excluding each object. Second, we use an iterative image inpainting algorithm that progressively accepts generated image regions with high fidelity. The advantage of such a model is being able to generate realistic content promptly to fulfill the requirements of real-time responses.

Several user studies demonstrate that our system provides flexible interfaces and accurate algorithms for clutter detection and removal, which allow users to identify visual distractions and take images of higher quality within less time. Users also report that the system helps them save time and explore more scenes and bolder compositions for better photographic work.

%% file: 2-Related.tex
\section{Related Works}

Our work lies at the intersection of image processing, photography guidance, and image aesthetic evaluation. In this section, we discuss related work in these three fields of study.

\subsection{Image Processing}

In our system, we remove detected clutter from the photo and generate semantically reasonable and visually plausible contents for the missing regions based on background texture. In the field of computer vision, \emph{image inpainting}~\cite{bertalmio2000image} is the technique that synthesizes alternative contents for the corrupted regions~\cite{kim2021learning,li2021faceinpainter,wang2021image,peng2021generating,liu2021pd,liao2021image}. This technique has been extensively studied in many image processing tasks such as image re-targeting, movie restoration, object removal, and old photo restoration~\cite{yu2021wavefill}.

Conventional image inpainting methods exploit neighborhood information for content refilling. Specifically, diffusion methods~\cite{bertalmio2000image,ballester2001filling} propagate neighboring information to the corrupted regions by methods such as interpolation. When the corrupted region is large, there is a lack of global information, and it is hard for diffusion methods to recreate meaningful structures. Patch-based methods~\cite{barnes2009patchmatch,darabi2012image} complete images by searching and reusing similar patches from the background. They work well for images with repetitive contents but struggle with generating meaningful semantics.

Deep learning motivates image inpainting research efforts to shift toward data-driven learning-based approaches~\cite{wu2020leed,wu2020cascade,yu2021diverse,zhan2021bi,zhan2021unbalanced,zhan2019spatial}. Particularly, generative adversarial networks (GANs)~\cite{goodfellow2014generative} have proven their ability to infer contents for corrupted regions with both reasonable structures and realistic appearance~\cite{pathak2016context}. \citet{nazeri2019edgeconnect} supplement GANs with edge detection and salient edge prediction. The idea is that edges provide predictable structural information for corrupted regions. \citet{wang2018image} introduce different receptive fields into image inpainting by utilizing a multi-column network structure. \citet{zeng2019learning} design a hierarchical pyramid-context encoder to use the information at multiple scales for recovering the missing regions. \citet{liu2020rethinking} disentangle structures and textures and recover them separately by deep and shallow features. \citet{liu2018image, yu2019free} propose partial and gated convolution for image inpainting, respectively. Although algorithms based on GANs significantly improve image inpainting performance, they often generate artifacts for large missing regions. \citet{zeng2020high} provide an iterative method to fill the missing regions progressively. At each iteration, they treat inferred pixels with a high confidence value as part of the original images. A drawback of such an iterative method is the relatively low inference speed. In this paper, we borrow the idea of iterative image inpainting and accelerate its speed for fulfilling the needs of real-time interactions.

Related to our work, \citet{fried2015finding} incrementally remove automatically detected distractors by dragging a slider. Our work differs from this system in (1) \citet{fried2015finding} focus on small visual elements while our method is effective for distractors of various sizes; (2) instead of training the distractor detection model using a supervised learning method as in \cite{fried2015finding}, our clutter detection method is based on aesthetic evaluation of objects, which is more suitable for photographic guidance; (3) we additionally develop a system that visualizes, interactively determines and removes clutter from photos in real-time.

\subsection{Capture-Time Guidance} In-camera guidance is a hot topic in human-computer interaction research. Previous work helps photographers by positioning their cameras for better compositions~\cite{bae2010computational,carter2010nudgecam,mitarai2013interactive,fried2020adaptive}, displaying view proposals~\cite{ma2019smarteye,rawat2015context}, guiding users to better lightning~\cite{jane2019optimizing,li2017guided}, and incorporating aesthetic evaluation models to provide suggestions on improving photos~\cite{wu2021tumera,wu2022interpretable}. Clutter is not the focus of these works.

Our work is closely related to \citet{jane2021dynamic}. In this paper, the authors also study the influence of clutter on photographs. They develop a system providing visual overlays highlighting the edges of objects. The overlays are expected to prompt the exploration of creative concepts and the understanding of the application of higher-level photography principles. Beyond reminding the photographers of the existence of clutter, we are interested in better identifying and guiding the users to get rid of the clutter. Therefore, we introduce a computational model and define clutter as those objects which contribute negatively to the aesthetic appearance and contents of the photo. We give suggestions about removing clutter. Furthermore, for those objects that can not be easily removed, we provide an image inpainting tool. These points make our system different from \cite{jane2021dynamic} with respect to functions, interactions, visualization, and underlying techniques.

\subsection{Aesthetic Evaluation}
Automatic image aesthetic assessments promote many real-world applications~\cite{westerman2007creative, wu2021tumera, wang2017deep, bhattacharya2010framework,datta2007learning} and human-computer interactions~\cite{tractinsky1997aesthetics,kurosu1995apparent, tullis1981evaluation,tullis1984predicting, kelster1983making,aspillaga1991screen,toh1998cognitive,szabo1999effects,heines1984screen,grabinger1993computer}. Conventional approaches rely on domain knowledge and design \emph{handcrafted features} for evaluation. Hand-designed features generally belong to four categories. (1) Basic visual features, including color distribution~\cite{ke2006design,aydin2014automated}, hue count~\cite{ke2006design}, blurriness~\cite{tong2004classification}, sharpness~\cite{aydin2014automated}, depth~\cite{aydin2014automated}, global edge distribution~\cite{ke2006design}, low-level contrast~\cite{tong2004classification,ke2006design}, and brightness~\cite{ke2006design}. These features can also be extracted locally from potentially attractive image regions~\cite{luo2008photo,wong2009saliency,nishiyama2011aesthetic,lo2012statistic}. (2) Composition features~\cite{bhattacharya2010framework,dhar2011high,bhattacharya2011holistic,wu2010good,wu2010good,tang2013content,zhang2014fusion} that are designed to model techniques like the rule of thirds, low depth of field, and opposing colors, whose purpose is to make the major object salient. (3) General-purpose features that are not specifically designed for aesthetics assessment, such as Fisher vector (FV)~\cite{marchesotti2011assessing,marchesotti2013learning,murray2012ava}, scale-invariant feature transform (SIFT)~\cite{yeh2012relative}, and bag of words (BOV). (4) Task-specific features that are designed specifically for a category of images. For example,~\citet{li2010aesthetic} and~\citet{lienhard2015low} design features for human faces, \citet{su2011scenic} and~\citet{yin2012assessing} focus on landscape photos, and \citet{sun2015aesthetic} evaluate the visual quality of Chinese calligraphy. Based on these features, heuristic rules~\cite{aydin2014automated}, SVMs~\cite{datta2006studying,luo2008photo,wong2009saliency,wu2010good,nishiyama2011aesthetic,dhar2011high,wu2010good,yin2012assessing,tang2013content,lienhard2015low}, boosting~\cite{tong2004classification,su2011scenic}, regression~\cite{sun2009photo}, SVR~\cite{bhattacharya2010framework,li2010aesthetic,bhattacharya2011holistic}, naive Bayes classifiers~\cite{ke2006design}, and Gaussian mixture model~\cite{marchesotti2011assessing,marchesotti2013learning,murray2012ava} are used to get the overall aesthetic score estimation. For a detailed discussion of these methods, we refer readers to~\citet{deng2017image}. 

Handcrafted features make good use of human knowledge of aesthetics, but their design requires a large number of engineering efforts and is not flexibly applicable to different aesthetics tasks. Due to these shortcomings, they largely underperform methods based on automatic feature extraction by learning deep neural networks.

\citet{wang2016multi} modify AlexNet for \emph{deep aesthetic evaluation}. Specifically, the authors replace the fifth convolutional layer of AlexNet with a group of seven convolutional layers, each of which is dedicated to one category of scenes. The outputs from these layers are averaged, and the whole architecture is trained end-to-end with backpropagation. \citet{tian2015query} train an aesthetic model with a smaller fully-connected layer as the feature extractor and an SVM as the classifier. Images are usually resized or cropped before being fed into the deep model because the model typically requires the input to be of the same dimensions. To prevent visual features from being corrupted during this process,~\citet{lu2015deep} randomly select image patches of the same size. These patches are directly fed into the deep model to retain the original aesthetic information. Other solutions include adaptive spatial pooling~\cite{mai2016composition}. To deal with images with different scenes, multi-column neural networks have also been used~\cite{mai2016composition}. This structure is also used for global-local feature combination~\cite{lu2015rating}, patch-global input tradeoff~\cite{zhang2016describing}, and separate object-scene-texture processing~\cite{kao2016hierarchical}.

\citet{kong2016photo} present an \emph{aesthetics dataset} providing an overall aesthetic score and eleven aesthetic attribute scores for over 10,000 images. This paper also proposes a Siamese architecture that estimates the relative aesthetic quality of two input images for better distinguishing images of similar quality. \citet{wang2016brain} predict different style attributes and use them as the input to a final CNN for predicting the overall score distribution. \citet{wu2022interpretable} push forward this line of research by introducing interpretability into deep aesthetic models. The author estimates the contribution of each attribute to the overall aesthetic score via learning a hyper-network and estimates the contribution of different image regions to attribute scores by introducing an attentional mechanism. Similar to \citet{wu2022interpretable}, we also use an attention model, but the model is specially designed to reveal the contribution of cluttered objects to the overall aesthetic appearance and contents. This method draws inspirations from cooperative multi-agent learning~\cite{wen2022multi,kuba2021trust,christianos2020shared,jiang2019graph,dong2022low,foerster2018counterfactual,lowe2017multi,dong2023symmetry,singh2019learning,qin2022multi,dong2021birds,rashid2020weighted,ivanov2024principal,hossain2024multi,foerster2016learning,zhang2024position}, where a decomposed value function is learned for credit assignment~\citep{rashid2018qmix,sunehag2018value,kang2022non,peng2021facmac,wu2021containerized,wang2025bandit}.

%% file: 3-Design.tex
\begin{figure}
    \centering
    \subfigure[Pictures of Group 3 taken before and after reading the critic's comments.]{\includegraphics[width=0.48\linewidth]{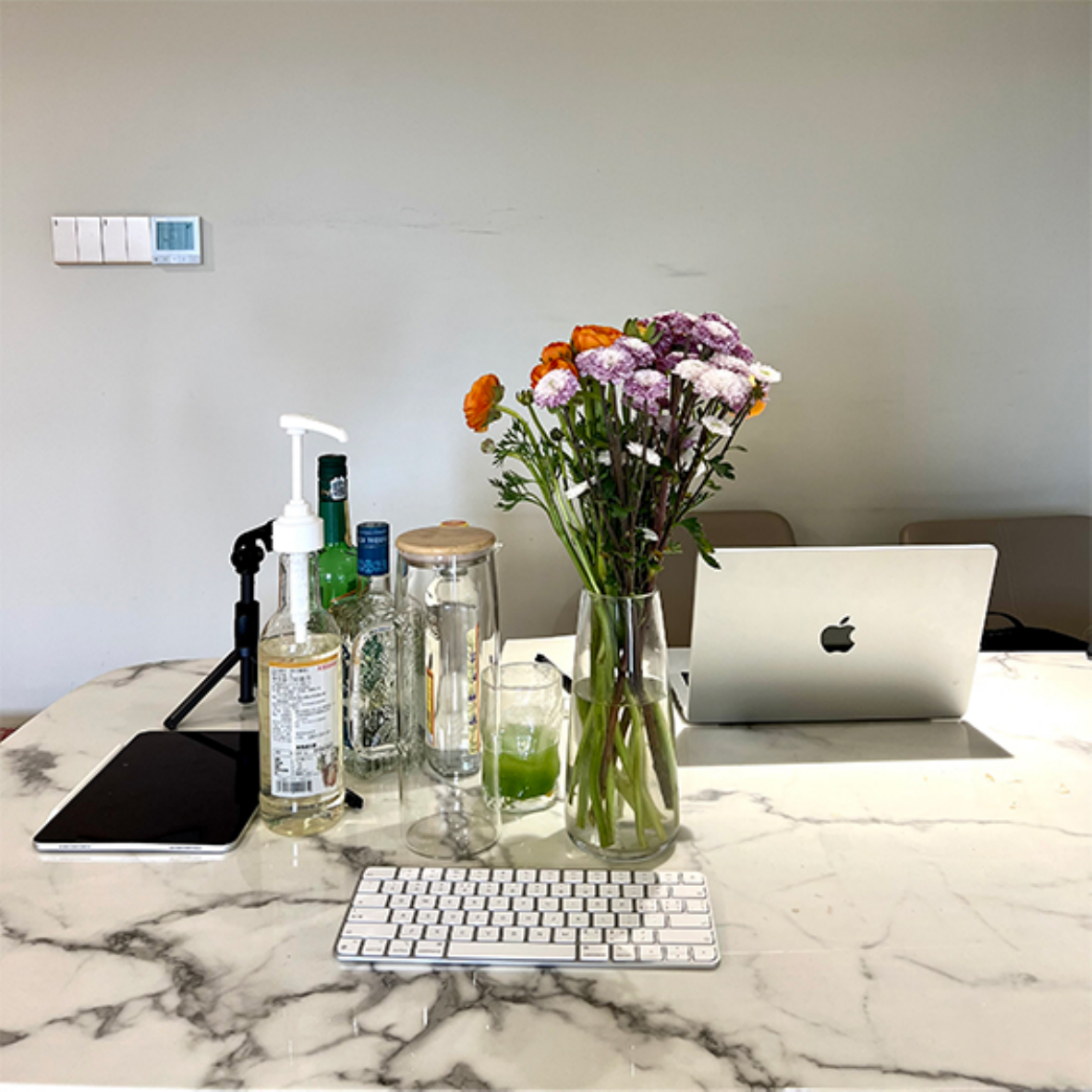}\hspace{0.04\linewidth}
    \includegraphics[width=0.48\linewidth]{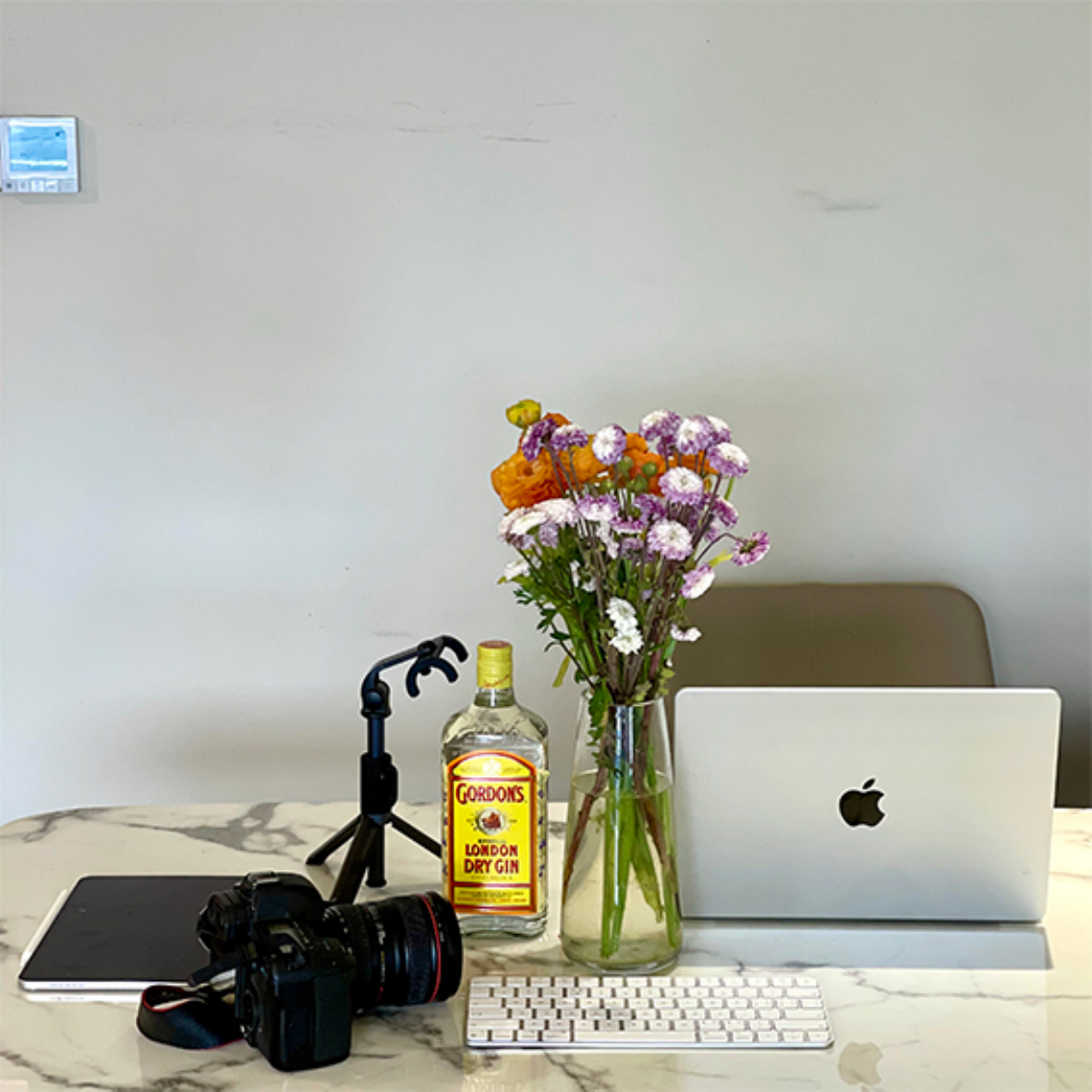}}
    \subfigure[Pictures of Group 6 taken before and after reading the critic's comments.]{\includegraphics[width=0.48\linewidth]{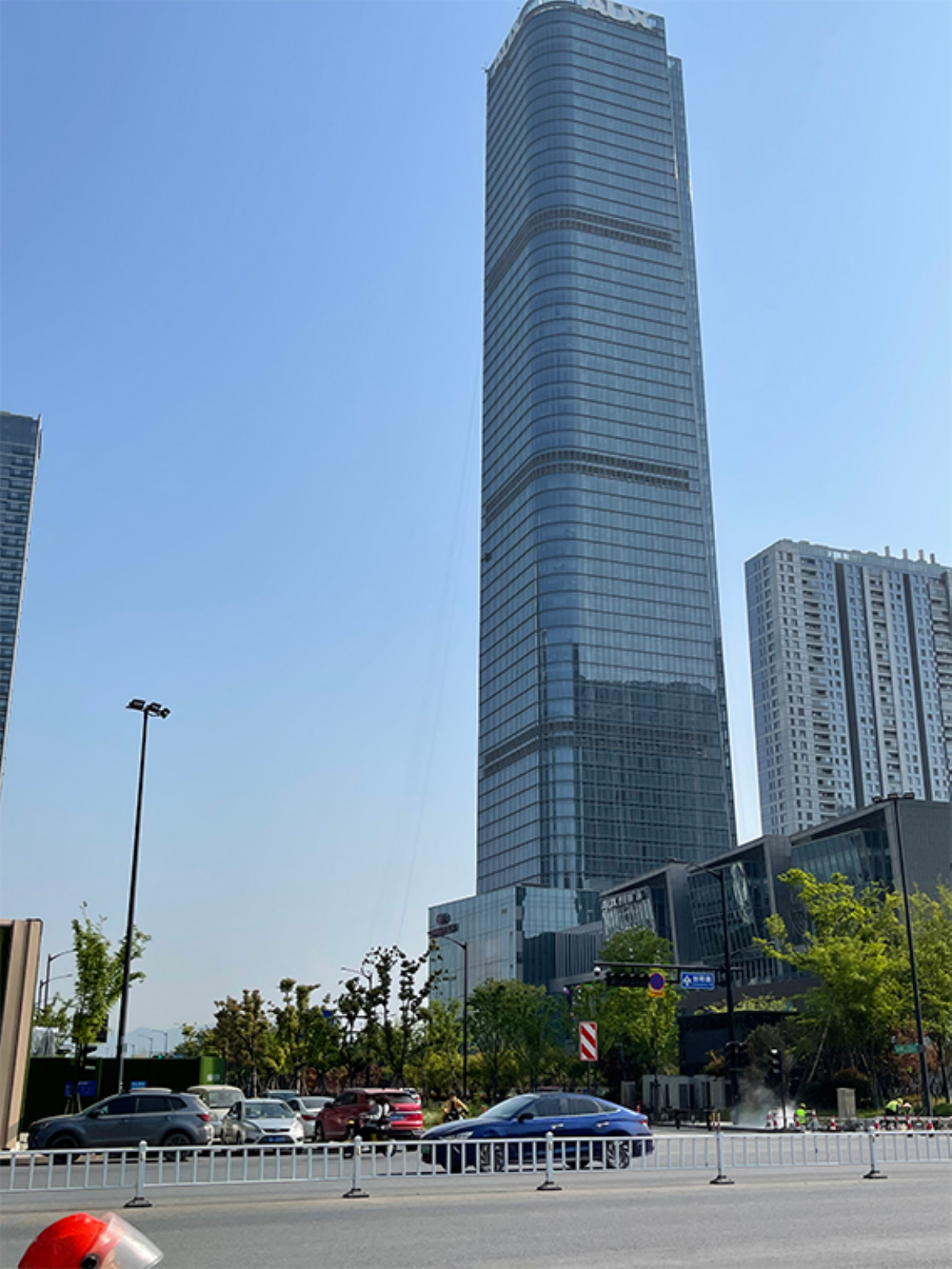}\hspace{0.04\linewidth}
    \includegraphics[width=0.48\linewidth]{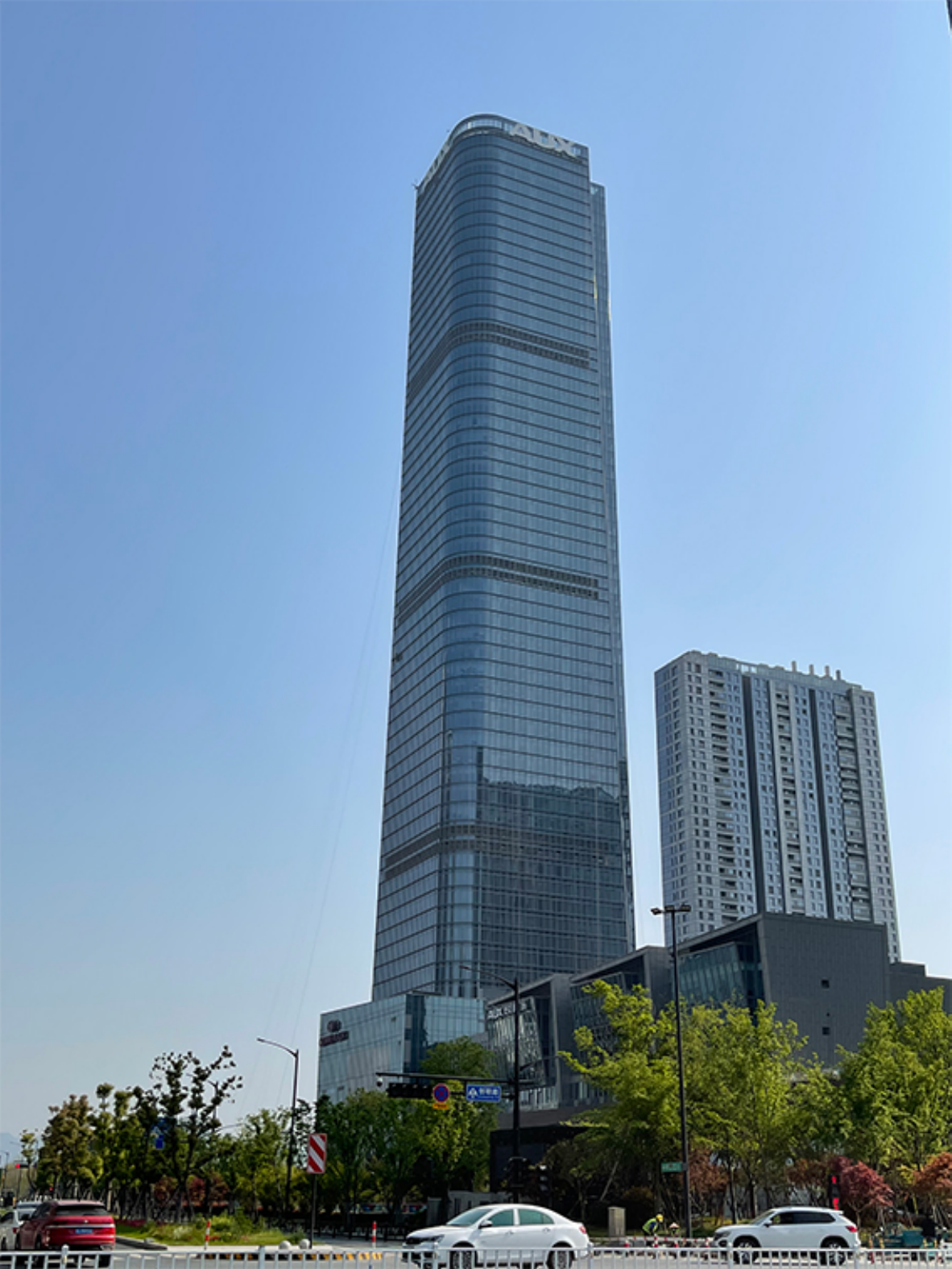}}
    \caption{Photos of Group 3 and 6 in our visual clutter survey.}
    \label{fig:visual_survey}
\end{figure}
\section{Design Process}\label{sec:design}

In this section, we describe our design process towards an interface that reminds the user of cluttered objects, suggests a removal scheme of these objects, and guides the user to take photographs of higher quality. We will first analyze the reasons why photographers occasionally include clutter in their photographic work. Then, we design the interfaces to effectively guide the user to deal with different cluttered objects. These interactions motivate us to develop novel AI methods which are introduced in Sec.~\ref{sec:method}.

Our design process aims to answer the following questions: (1) What kinds of clutter can distract the audience? (2) How to remind the photographer of different kinds of clutter? (3) How to help users remove the clutter to improve photo quality?

\subsection{Visual Clutter Survey}\label{sec:survey}

We start with a survey about how visual elements affect users' perception of a photo. The aim of this survey is to identify different kinds of clutter and get insights into how to handle them. 

\subsubsection{Setup} Our survey has 24 participants, with an equal number of females and males. They are aged 23 to 39 years old, with an average age of 28. We didn't provide compensation, and the participation was voluntary. The participants are randomly divided into 8 groups, with 3 experimenters in a group. 

The survey goes in the following three steps. (1) Photography. For each group, one of the participants (the photographer) is asked to take a photo with the default camera app of the phone. This participant is then asked to give a caption describing the intended story. (2) Feedback. The second experimenter in the group (the critic) is tasked to write down its opinion about the photo. Specifically, without knowing the caption of the photographer, the critic tells the story of the photo according to what it sees. The critic is also asked which visual elements support its story. (3) Re-photo. The photographer reads the critic's comments. If the story is significantly different, the photographer can take another photo telling the same intended story. The third experimenter in the group (the mediator) sees the two photographs and comments on the stories. 

\subsubsection{Results and Analyses} 

For the first example, the photographer in Group $3$ shoots the scene of her home office, including electronic devices and auxiliary equipment for remote working use, as well as daily necessities such as wine and bottles (Fig.~\ref{fig:visual_survey} (a) left). The photographer aims to document "\emph{the special experience that closely combines life and work during the epidemic.}" However, the critic tells a different story: "\emph{My attention was first drawn to the wine, syrup, and bottles in the middle of the scene. Clearly, there is a bottle of mixed drinks there. Then I noticed the iPad and the laptop on the table. The flowers should be for decorative purposes. Based on these observations, I guess the photographer wants to share that she was learning to stir up mixed drinks online}."

Here we see how the clutter prevents the convey of the photographer's intentions. Wine, bottles, and syrup take up a large part of the picture. The photographer thinks that these objects are not so obtrusive because she works here and is accustomed to these objects. By contrast, from the eyes of the critic, these objects together draw his attention, and he thinks these objects are the major part of the photo. We see $\mathtt{the\ first\ kind\ of\ clutter}$ in this case: the photographer gets used to some objects and is not aware that they draw much attention and clutter the picture. 

On receiving the feedback from the critic, the photographer becomes more intentional about staging the elements that are key to her stories. She removes many bottles from the scene and zooms in to emphasize the home scene (Fig.~\ref{fig:visual_survey} (a) right). We find that, in order to get rid of clutter, many (4) photographers choose to remove objects according to the critic's comments, while others choose to zoom in (2), adjust the lighting of the scene (2), or change orientation from portrait to landscape (1).

The mediator successfully gets the intention of the photographer and guesses that she is framing a home working scene. However, the mediator was distracted by some other visual elements when he saw the photo for the first time: "\emph{My eyes stopped on the light switch for a long time. I know it was not the critical part, but I just cannot help myself. I started to see other parts of the photo only after I could tell the symbol on the switch}." The light switch is a representative example of $\mathtt{the\ second\ kind\ of\ clutter}$ -- the photographer is focusing on other aspects of the photo and ignores some visual elements that are irrelevant to the story of a photo but are obtrusive.

For the second example, the photographer of Group $6$ (Fig.~\ref{fig:visual_survey} (b) left) frames a street scene with a magnificent building to express "\emph{the joy of getting a new job in a beautiful building}." However, the critic thinks: "\emph{I guess the photographer is documenting a street scene for the building, but the vehicles and street lights clutter the photo, some of which are not visually pleasant. Elements like these take the attention away from the subject of the photo.}" Based on the comments of the critic, the photographer seeks to get rid of the clutter in the second attempt. However, he finds it difficult because the traffic is heavy and there is a construction site that is very close to the building. Therefore, as shown in Fig.~\ref{fig:visual_survey} (b)-right, the second attempt is still cluttered with visual distractions. The mediator says: "\emph{From the comparison of these two photos, the photographer seeks to reduce clutter in the scene. It is not easy. However, this situation is quite common -- just imagine you are taking photos at a resort with many tourists.}" The mediator's comments reveal $\mathtt{a\ third\ kind\ of\ clutter}$: the photographer is aware of the clutter, but it is hard to remove it from the scene.

\begin{figure*}
    \centering
    \includegraphics[width=0.95\linewidth]{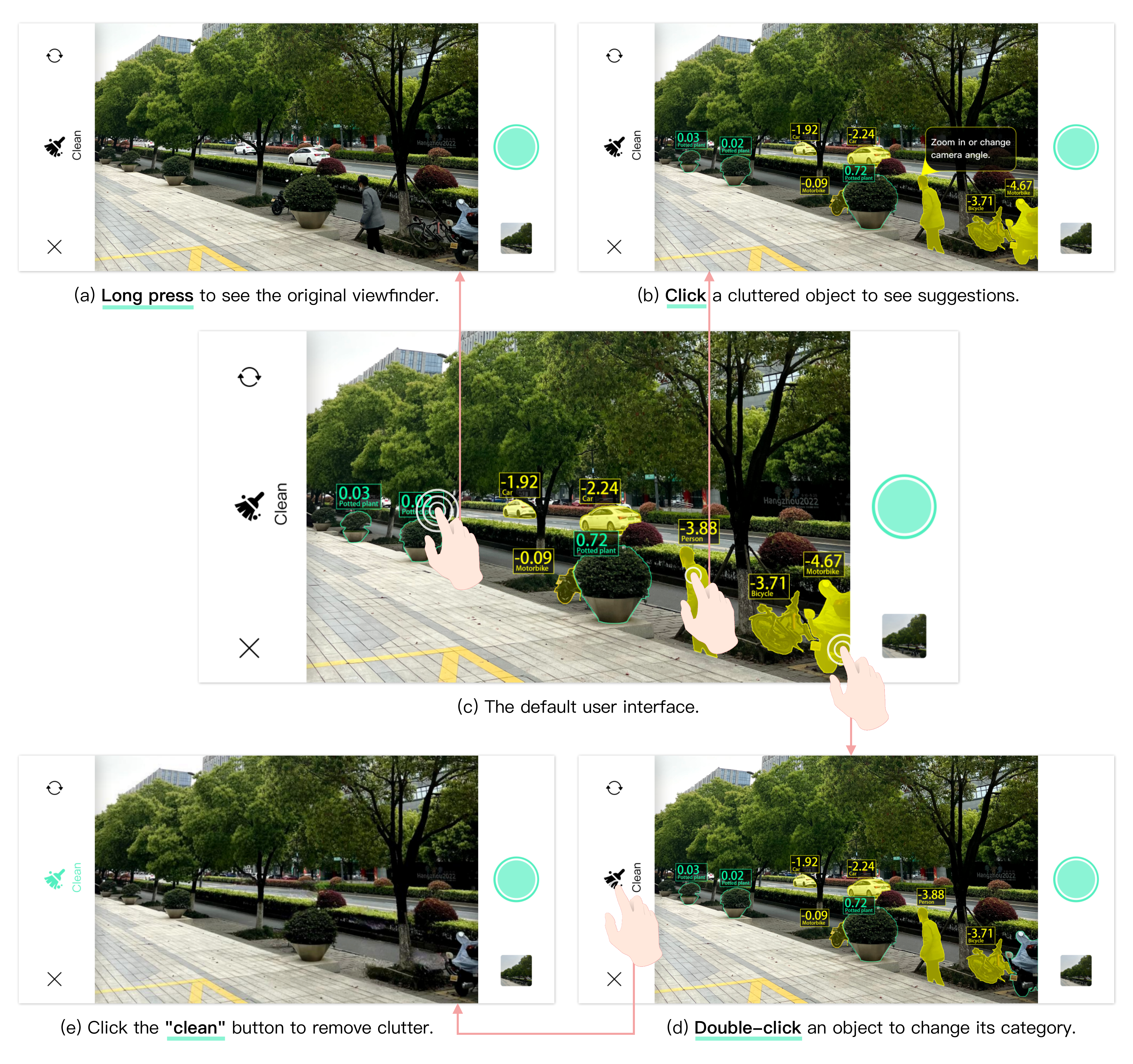}
    \caption{Interfaces and interactions of our system. On the default interface \textbf{(c)}, detected clutter and other objects are highlighted by different masks. Shown above a mask is the estimated contribution value to the overall aesthetic appealingness and content. By pressing the screen, users see the original image in the viewfinder \textbf{(a)}. Suggestions about how to deal with a cluttered object \textbf{(b)} can be obtained by clicking the object. Users can change the category of an object by double-clicking it \textbf{(d)}. We also provide a "clean" button that removes current clutter in the scene \textbf{(e)}.}
    \label{fig:interaction}
\end{figure*}

\subsubsection{Survey Insight} The two examples discussed in the previous sub-section cover the three kinds of clutter that we find in our survey: (1) The photographer gets accustomed to some visual elements, but these elements are obtrusive to the audience; (2) The photographer ignores some visual distractions when focusing on other aspects of the photo; (3) The photographer knows the clutter, but it is difficult to remove it. We focus on how to design interactions and develop algorithms to help the user get rid of these three kinds of clutter in this paper. 

The first kind of clutter is challenging to handle because the gap between this kind of clutter and meaningful content is narrow. We propose to quantify the contribution of objects to the overall aesthetic perception and content to distinguish clutter and meaningful content. Dealing with the second kind of clutter is relatively easy. We can highlight the neglected clutter to remind the photographer of it. For the third kind of clutter, we propose to remove the cluttered objects by developing a real-time image inpainting algorithm that fills the regions of clutter with reasonable background content.

In summary, to help users avoid clutter in photos, we need algorithms to detect objects in an image, quantify the contribution of objects to the overall content and aesthetic quality, and remove selected objects. We also need a user interface to visualize the computational results, remind the user of clutter, and provide flexible interactions for clutter detection and removal. In Sec.~\ref{sec:method}, we will introduce our technical novelties for realizing these functions. Before that, we first describe how we build an effective user interface that helps clutter avoidance in a user-friendly manner.


\subsection{Interaction Design}

Our system is a camera app with standard photographing functions and several novel interactions related to clutter removal, which are highlighted in bold in the following paragraphs.

On opening our system, the user sees the viewfinder with \textbf{masks highlighting} (Fig.~\ref{fig:interaction} (c)) \textbf{objects} in the scene. At the top of a mask are a label showing the name of the corresponding object and \textbf{a number} ($q$) \textbf{indicating\ the\ contribution\ of\ this\ object} to the overall content and aesthetic quality. The contribution $q$ can be positive or negative, with a negative value indicating that the object lowers the overall quality and may be clutter. We classify objects with negative $q$ values as clutter. While normal objects are highlighted with a transparent mask, cluttered objects are emphasized with a bright color. These masks and contribution values can help deal with the first two kinds of clutter discussed in Sec.~\ref{sec:survey} -- the user gets aware of the ignored cluttered objects. If the user wants to see the original scene, \textbf{the masks become invisible when the user presses and holds the viewport} (Fig.~\ref{fig:interaction} (a)).

\textbf{By\ clicking\ an\ object\ mask,\ the\ user\ gets\ suggestions\ about how to\ get\ rid\ of\ it} (Fig.~\ref{fig:interaction} (b)). For clutter around the photo boundary, the suggestions include zooming in, changing the orientation from portrait to landscape, and adjusting the camera angle. For clutter in the middle part of the image, there are two suggestions: moving it out of the scene by hand if it is convenient or using our image inpainting algorithm to remove it. The user can follow these suggestions to improve the photographic work.

A question about our classification algorithm of normal and cluttered objects is that the results may conflict with the user's judgment. Therefore, we provide an interface where the user can change the classification of an object. \textbf{By\ double-clicking\ an\ object\ mask,\ the user\ flips the classification of the corresponding object} (Fig.~\ref{fig:interaction} (d)). For example, a normal object will be regarded as a cluttered object after being double-clicked, and vice versa. The color of the mask will also change immediately to reflect this change. 

According to the suggestions, the user takes possible actions to remove some of the cluttered objects. For those objects that are not easy to be removed, such as other pedestrians and vehicles, \textbf{the user can click the "clean" button to trigger the image inpainting algorithm and preview the scene without these objects} (Fig.~\ref{fig:interaction} (e)). On clicking this button, the back end will remove the objects that are currently classified as clutter. At capture time, for real-time response, we run a low-fidelity image inpainting algorithm. If the user likes the photo, it can click the shoot button and save the original image in the viewfinder. The user can choose to clean up the cluttered objects for the save photos by clicking the "clean" button in the album to call a high-fidelity image inpainting algorithm.

These interactions are based on the object masks, the estimated contribution value $q$, and the inpainted images with cluttered objects removed. In the next section, we describe in detail how to develop algorithms that provide the information.

%% file: 3-Method.tex
\section{Computational Methods}\label{sec:method}

The results of our survey and the realization of our interaction interfaces described in Sec.~\ref{sec:design} motivate us to develop computational modules for the following three functions: (1) detecting objects from the scene; (2) classifying normal and cluttered objects by calculating their contribution to the overall content and aesthetic quality; (3) removing the cluttered objects and reconstructing the missing regions with realistic background context. In this section, we describe the technical details of the realization of these functions.

The first function, object detection, is extensively studied in the field of computer vision. Mask R-CNN~\cite{he2017mask} is a classic, lightweight, and accurate deep model that predicts the object labels, gives bounding boxes, and provides pixel-level masks showing the detailed position of objects (instance segmentation). Formally, given an input image of the size $h\times w$, we use the instance segmentation head of a trained Mask R-CNN model that outputs $k$ masks $\{m_1, m_2,\dots,m_k\}$ for $k$ detected objects. Each mask $m_i\in\mathbb{R}^{h\times w}$ has the same size as the input image. The values in masks are binary, with $1$ meaning that the corresponding pixel belongs to the object and $0$ meaning the opposite. These masks will be used to highlight objects on the screen, as well as in other modules of our computational system.

For the other two functions, there are no off-shelf methods that meet our needs. We develop algorithms to realize these functions, which are described in the next two sub-sections.

\begin{figure}
    \centering
    \includegraphics[width=\linewidth]{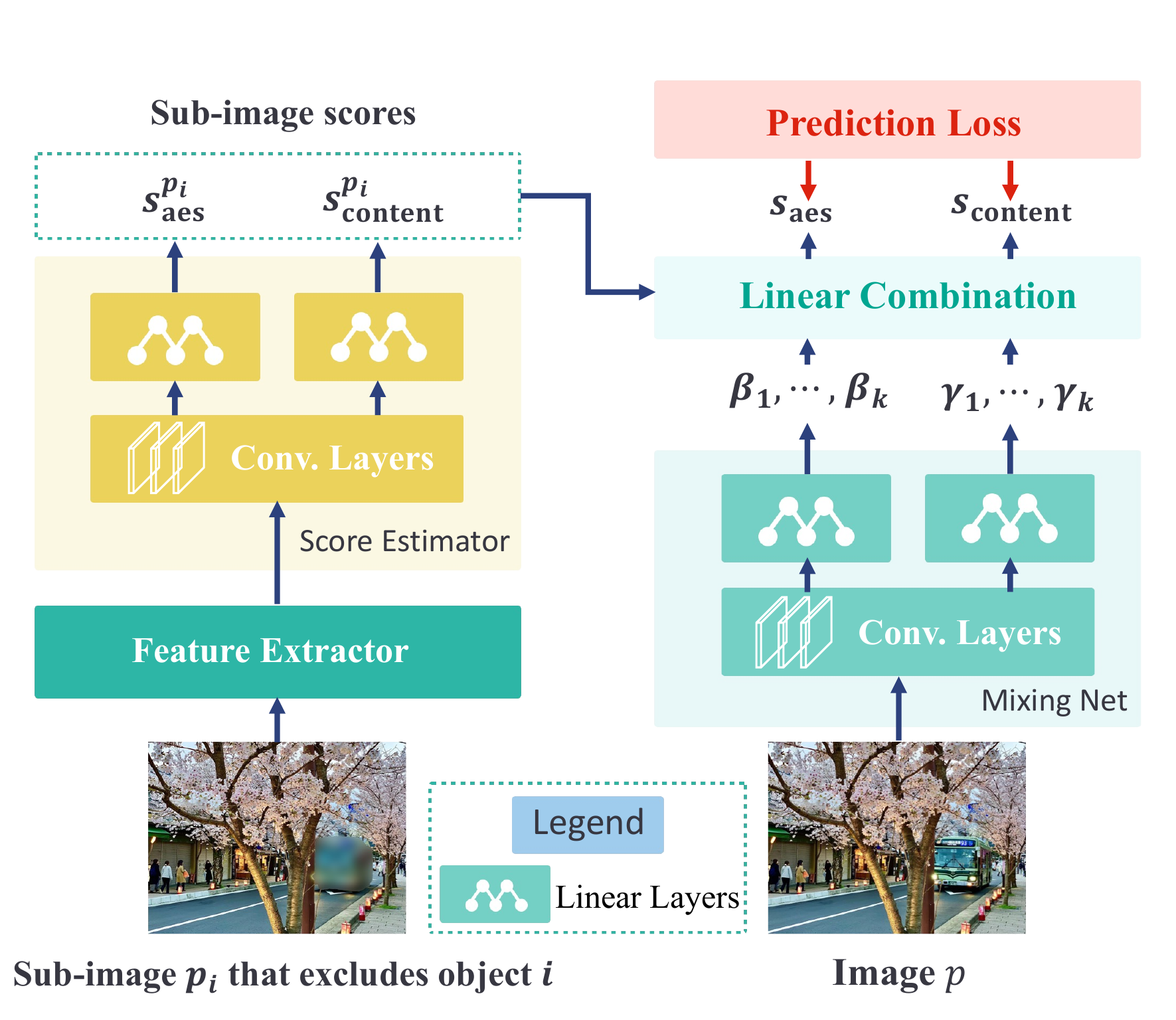}
    \caption{Network architecture of our framework that estimates the contribution of objects to the overall aesthetic quality and content. }
    \label{fig:aes-framework}
\end{figure}

\subsection{Clutter Distinguishment and Contribution Quantification}\label{sec:aes}

In this sub-section, we introduce a novel deep neural network model that learns the contribution of each object to the overall content and aesthetic quality and an approach to distinguish clutter from normal objects based on this model.

Modern image aesthetic datasets, like the AADB~\cite{kong2016photo} dataset, provide scores, $s_{\text{aes}}$ and $s_{\text{content}}$, for each image reflecting whether the image is aesthetically appealing and contains interesting content, respectively. Our core idea here is to learn these overall scores as a linear combination of scores for sub-images that exclude given objects. Based on this linear combination, we can calculate the contribution of each object.

The proposed learning framework (Fig.~\ref{fig:aes-framework}) consists of three modules. The first part is a convolutional neural network (CNN) as a feature extractor. We use the AADB dataset for learning and have a limited number (9870) of training samples with human-labeled scores. Therefore, instead of training the feature extractor from scratch, which is impractical, we use a pre-trained ResNet~\cite{he2016deep} for its proven ability of robust image representation learning that enables ever-improving performance in the task of aesthetic evaluation~\cite{mccormack2020understanding,jang2021analysis,mccormack2021deep}. Input to this feature extractor is a sub-image that blurs a given object. Specifically, for each object mask $m_i$ given by the object detection algorithm, we generate a sub-image 
$p_i$ by performing Gaussian blurring to the regions where the value of $m_i$ is 1. The blur is processed by calculating the convolution of image patches with a $13\times 13$ Gaussian kernel with a variance of 1. Sub-images for each object share a single feature extractor, which outputs feature maps, $z_{p_i}$, for them.

Feature maps $z_{p_i}$ are then fed into a score estimator, which outputs two scores for each sub-image. The first score $s^{p_i}_{\text{aes}}$ is an estimation of the aesthetic quality of the input, while the second score $s^{p_i}_{\text{content}}$ estimates the quality of content. The score estimator is a convolutional neural network with learnable parameters $\theta_{\text{score}}$. This network has two convolutional layers, a flatten layer, and two output heads. Each output head is a two-layer fully-connected network. We denote the score estimator by $f_{\text{score}}$, and $(s^{p_i}_{\text{aes}}, s^{p_i}_{\text{content}}) = f_{\text{score}}(p_i; \theta_{\text{score}})$.

To learn how much each object contributes to the overall score, we propose to learn the overall score as a combination of individual scores. Specifically, we learn
\begin{equation}
    s_{\text{aes}} = \sum_{i=1}^k \beta_i * s^{p_i}_{\text{aes}},
\end{equation}
where $s_{\text{aes}}$ is the overall aesthetic score prediction. Similarly, the overall content score is decomposed as 
\begin{equation}
    s_{\text{content}} = \sum_{i=1}^k \gamma_i * s^{p_i}_{\text{content}}.
\end{equation}
In this way, the overall score is represented as a linear combination of object scores. One question is how we can get the weights, $\beta$ and $\gamma$, of these linear combinations. Intuitively, the interrelationship between objects is determined by the structure of the original image. Therefore, we adopt a network conditioned on the input image to learn the linear combination parameters. Specifically, a network $f_{\text{mix}}(\ \cdot\ ; \theta_{mix})$ with two output heads, each with a SoftMax activation after the last layer, processes the input image and outputs the vector $\bm \beta=\langle \beta_1, \beta_2, \dots, \beta_k\rangle$ and the vector $\bm \gamma=\langle \gamma, \gamma_2, \dots, \gamma_k\rangle$.

It is worth noting that gradients can flow through $\bm \beta$ and $\bm\gamma$ into $f_{\text{mix}}$, so that its parameters can be updated. On the other hand, gradients can also flow into the score estimator through $s_{\text{aes}}$ and $s_{\text{content}}$. Therefore, our learning framework as a whole is end-to-end differentiable and can be learned by minimizing prediction losses.

We use two prediction losses to train our model. The first one is to minimize the mean square error between the predicted aesthetic quality $s_{\text{aes}}$ and the ground-truth $y_{\text{aes}}$:
\begin{equation}
    \mathcal{L}_{\text{aes}}(\theta_{\text{score}}, \theta_{\text{mix}}) = \mathbb{E}_{p\sim\mathcal{T}}\left[\left(y_{\text{aes}}(p) - s_{\text{aes}}(p;\theta_{\text{score}}, \theta_{\text{mix}})\right)^2 \right],\label{equ:loss_aes}
\end{equation}
where the expectation operator means that images $p$ are sampled from the training set $\mathcal{T}$. The second loss is for minimizing the content score prediction error that is given by:
\begin{align}
    \mathcal{L}_{\text{content}}&(\theta_{\text{score}}, \theta_{\text{mix}})\\ & = \mathbb{E}_{p\sim\mathcal{T}}\left[\left( y_{\text{content}}(p) - s_{\text{content}}(p;\theta_{\text{score}}, \theta_{\text{mix}})\right)^2\nonumber \right],\label{equ:loss_content}
\end{align}
where $y_{\text{content}}$ is the groundtruth score for the content provided by the dataset. By introducing a scaling factor $\lambda_{\text{aes}}$, the total loss for training our model is: $\mathcal{L}(\theta_{\text{score}}, \theta_{\text{mix}}) = \lambda_{\text{aes}} \mathcal{L}_{\text{aes}}(\theta_{\text{score}}, \theta_{\text{mix}}) + \mathcal{L}_{\text{content}}(\theta_{\text{score}}, \theta_{\text{mix}}). \nonumber \label{equ:total_loss}$

\subsubsection{Classify normal and cluttered objects}

With a trained score estimator, we can now distinguish cluttered objects from normal ones. Specifically, for the $i$th object in the scene, we obtain its aesthetic score $s^{p_i}_{\text{aes}}$, content score $s^{p_i}_{\text{content}}$, and the corresponding weights, $\beta_i$ and $\gamma_i$, by running our model. The semantic meaning of these scores is the quality of the image without the $i$th object. With the estimated overall scores for the original image, $s_{\text{aes}}$ and $s_{\text{content}}$, we can calculate the contribution of the $i$th object to the whole image as:
\begin{align}
    q_i = \beta_i(s_{\text{aes}} - s^{p_i}_{\text{aes}}) + \gamma_i(s_{\text{content}} - s^{p_i}_{\text{content}}).
\end{align}
A negative value of $q_i$ indicates that the $i$th object lowers the overall quality of the photo. Therefore, we classify these objects as clutter and the others as normal objects. If the classification conflicts with the impression of the user, it can change the category of an object by double-clicking its mask on the scene.

\subsection{Clutter Removal}\label{sec:removal}
The third function that we need to support is to remove the cluttered objects and fill the missing regions with realistic background context. The cluttered objects may be detected by our algorithms or selected by the user, and we can determine their positions by their masks $\langle m_{c_1}, \dots, m_{c_n} \rangle$, where $c_1, \dots, c_n$ are the indices of the cluttered objects and are in the set $\{1, \dots, k\}$. We assume that there are $n (\le k)$ cluttered objects in the photo.

As discussed in the related work section, image inpainting is the technique that replaces an object with background content. Generative adversarial networks (GANs) are frequently used to generate visually realistic content. These works use a generator to fill missing regions and a discriminator to distinguish generated images from real images. The generator is optimized to create images that are realistic enough so that the discriminator cannot distinguish. Although GANs facilitate significant performance improvements, running GAN-based image painting algorithms is time-consuming due to the large size of the model, and directly using them prolongs the response time of our system.

To reduce the response time, we use a small generative model. A consequence is that a small model generates artifacts that are not visually plausible for large missing regions. We adopt an iterative image inpainting approach to deal with artifacts. An additional, lightweight branch predicting which generated pixel is more like an artifact is added. In each iteration, we fill the part of the missing region with high-quality pixels and treat the other parts as a new missing region and run the next iteration. We note that~\citet{zeng2020high} use a similar iterative approach and find that it can improve the image quality in filled regions. Our method can be regarded as a lightweight and faster version of~\cite{zeng2020high}.

The input to the model is the image with a clutter mask $m_c$. We denote this corrupted image as $p_c=p \circ (1-m_c)$, where $\circ$ is the element-wise multiplication. The first branch of the generator $G$, parameterized by $\theta_g$, outputs the inpainted image: $y=G(p_c)$, while the second branch, parameterized by $\theta_{b}$, gives a probability map $b$ indicating how likely each pixel is an artifact. The generated image is trained by a reconstruction loss and an adversarial loss:
\begin{align}
    \mathcal{L}_G(\theta_g) = \mathbb{E}_{p\sim\mathcal{T}}\left[\|y-p\|_1 + (1-D(p\circ(1-m_c) + y \circ m_c))\right].
\end{align}
Here, $\mathcal{T}$ is the dataset for training, $\|y-p\|_1$ is the reconstruction loss, and $D$ is the discriminator parameterized by $\theta_d$. ($D$ has two outputs, with $1$ meaning a real image and $0$ meaning a fake image.) The generator is optimized to enforce the output of $D$ to be 1, meaning that the discriminator thinks $y$ is very likely to be a real image. In the meantime, the discriminator is trained to minimize
\begin{align}
    \mathcal{L}_D(\theta_d) & = \mathbb{E}_{p\sim\mathcal{T}}\left[1 - D(p)\right] \\ &+ \mathbb{E}_{p\sim\mathcal{T}, y=G(p)}\left[1 + D(p\circ (1-m_c) + y \circ m_c)\right],\nonumber
\end{align}
to be able to distinguish the generated images from the real images in the training dataset. The second branch of the generator is updated by minimizing:
\begin{align}
    \mathcal{L}_b(\theta_b) = \mathbb{E}_{p\sim\mathcal{T}}\left[ m_c \circ((1-b)\circ |y-p|)\right] .
\end{align}
By minimizing this loss, $b$ value is high only when the generated pixel in the missing regions is far away from the groundtruth pixel. Then we select regions with low $b$ values as the missing regions for the next iteration. To guarantee that our system responds on time, we set the maximum iteration time to 3 for capture-time guidance. After a photo is saved, the user can run a high-fidelity version of this algorithm by setting the maximum iteration time to 10. The back end would return the inpainted image if all $b$ values are lower than a threshold or the maximum iteration time is reached.

%% file: 5-Experiments.tex
\begin{figure}
    \centering
    \subfigure[$P15$'s viewfinder in the user interface.]{\includegraphics[width=\linewidth]{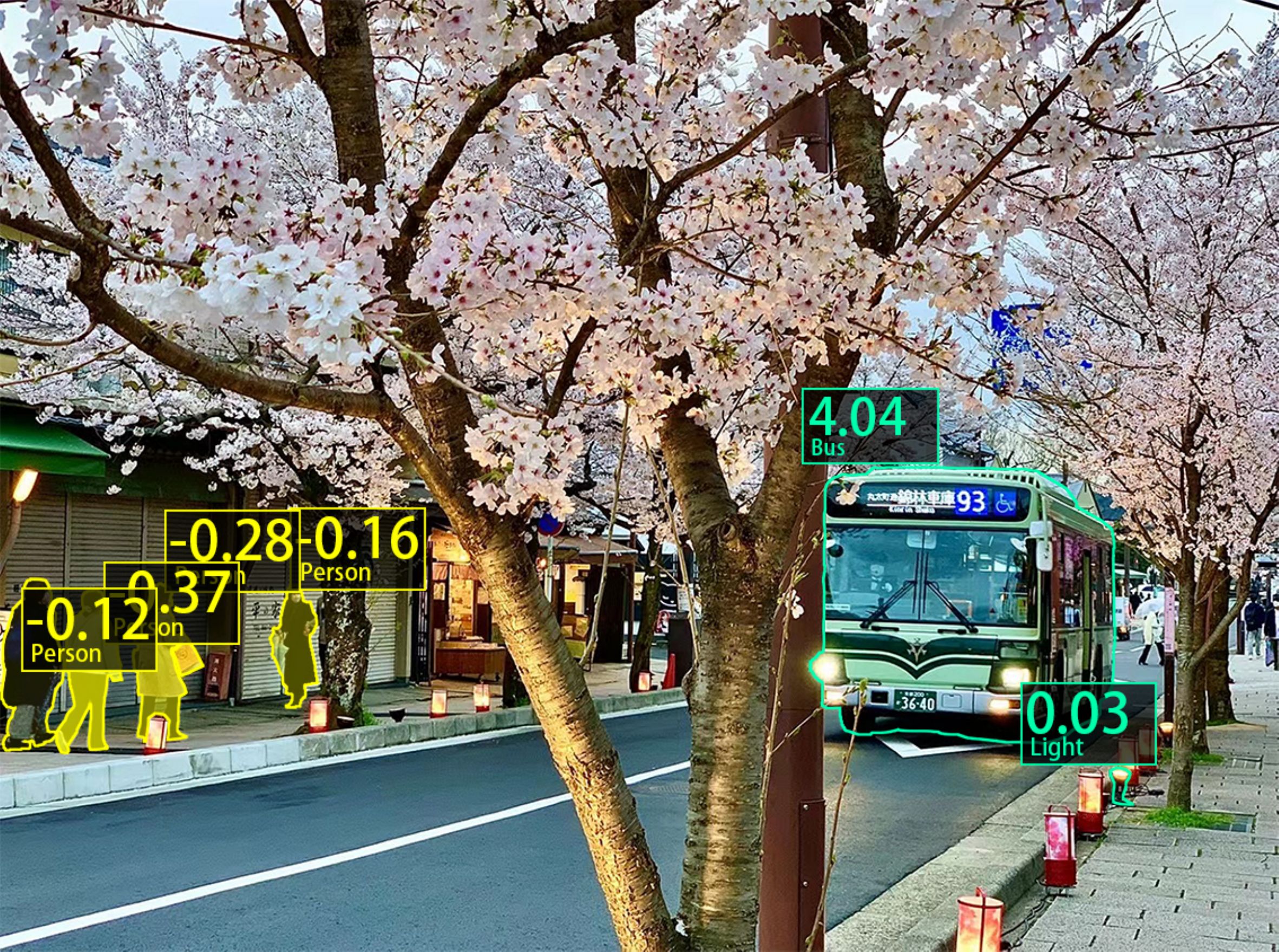}}
    \subfigure[$P4$'s viewfinder in the user interface.]{\includegraphics[width=0.49\linewidth]{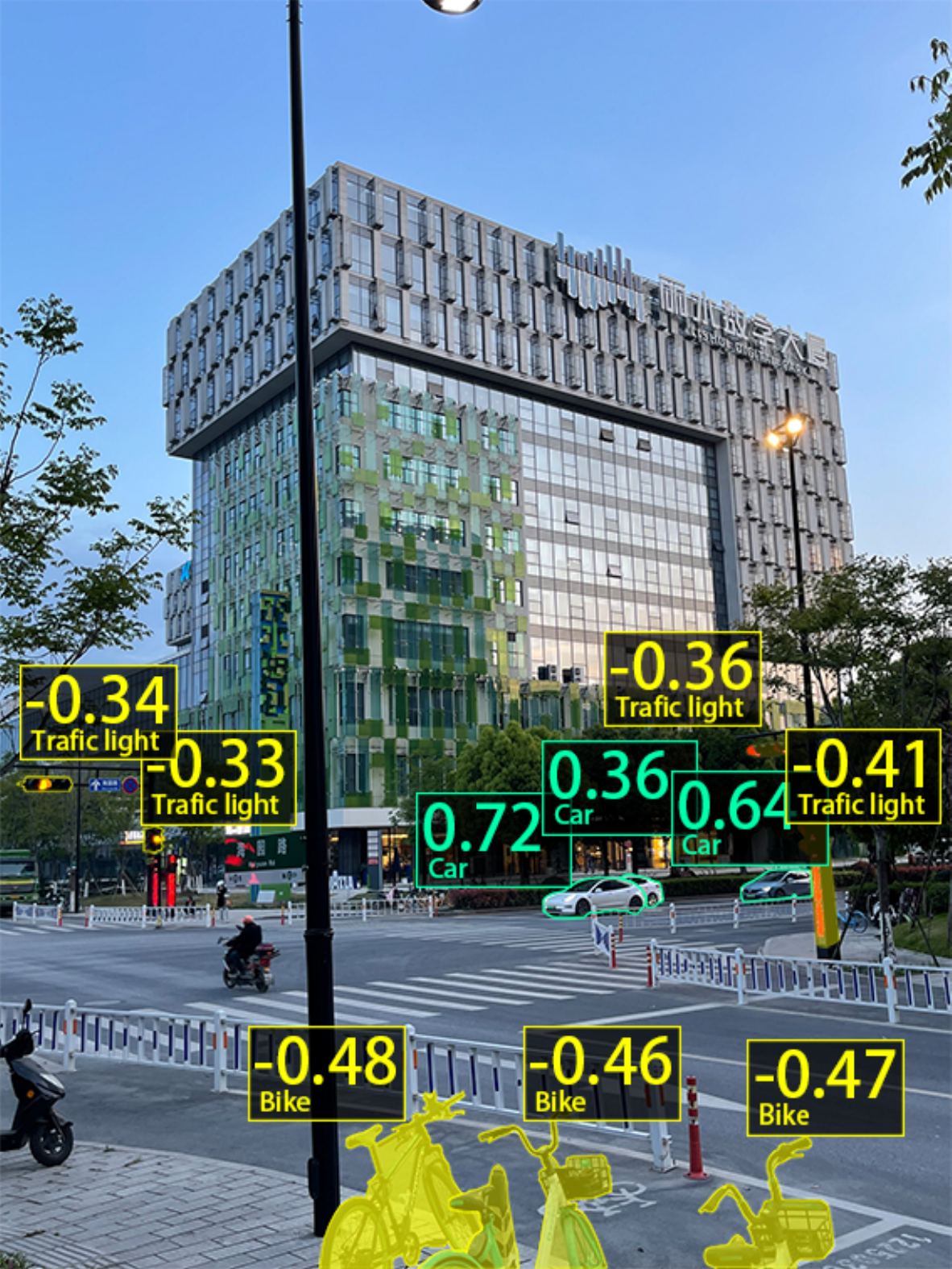}}\hfill
    \subfigure[$P1$'s viewfinder in the user interface.]{\includegraphics[width=0.49\linewidth]{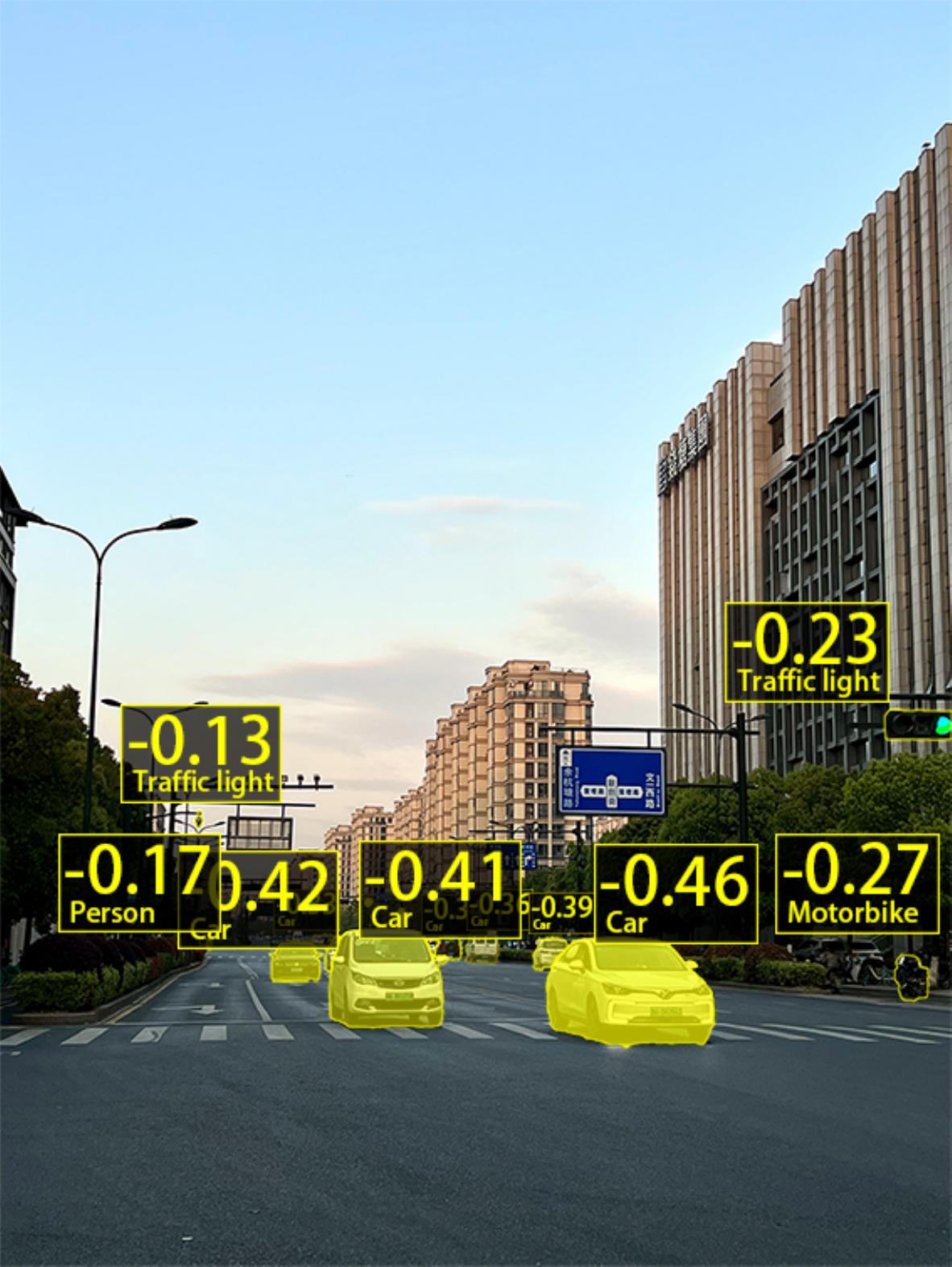}}
    \caption{Three examples of clutter detection. (a) $P9$ finds our system successfully recognizes the salient object and gives reasonable contribution estimations. (b) $P4$ also finds the detection results satisfactory. (c) Similar detection results frustrate $P1$ because they conflict with his intention.}
    \label{fig:detection}
\end{figure}

\section{Evaluation}\label{sec:exp-com}

In this section, we carry out user studies to study how users react to our system. Specifically, we are interested in (1) whether users think our clutter detection and removal algorithms are accurate; (2) whether our system can help users remove clutter and get better photographic works. Therefore, our user study consists of two major parts. The first part is for testing functional sub-modules of clutter detection and removal, and the second part is for the system as a whole. Participants of both parts of the study were compensated \$5.

\subsection{Study Procedure}
In this section, we will first introduce the details of the training scheme of our computational models and then describe the settings of user studies.

\subsubsection{Training Scheme}

We train our clutter classification model on the public image aesthetic dataset AADB~\cite{kong2016photo}. This dataset includes 10,000 images, and, for each image, we use its overall aesthetic quality score and its attribute score evaluating whether the content is interesting. These two scores are provided by five different human raters. Original images presented in this dataset are of different sizes. To fit our model, we resize the images to $256\times 256\times 3$. 

\begin{figure}
    \centering
    \subfigure[$P18$'s viewfinder in the user interface.]{\includegraphics[width=\linewidth]{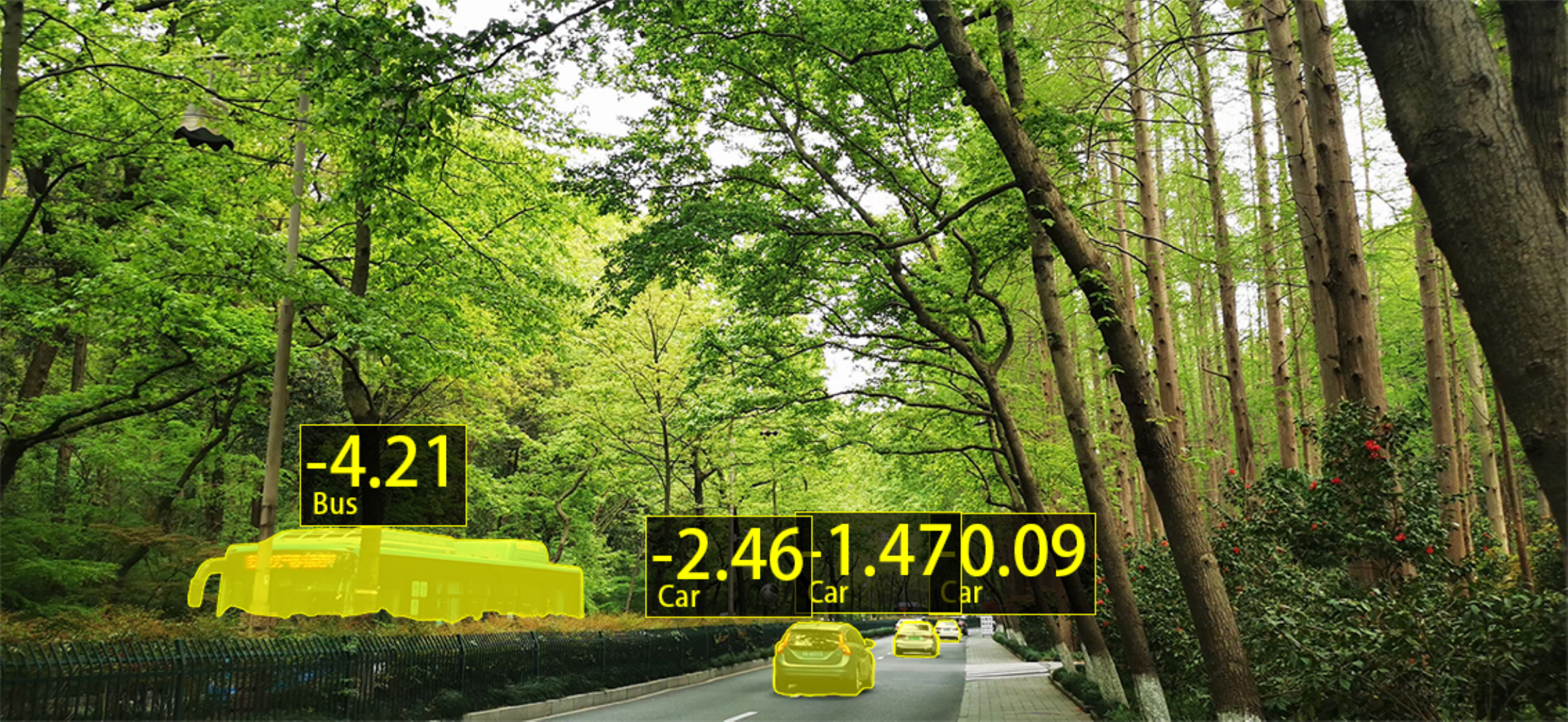}}
    \subfigure[$P11$'s viewfinder in the user interface.]{\includegraphics[width=\linewidth]{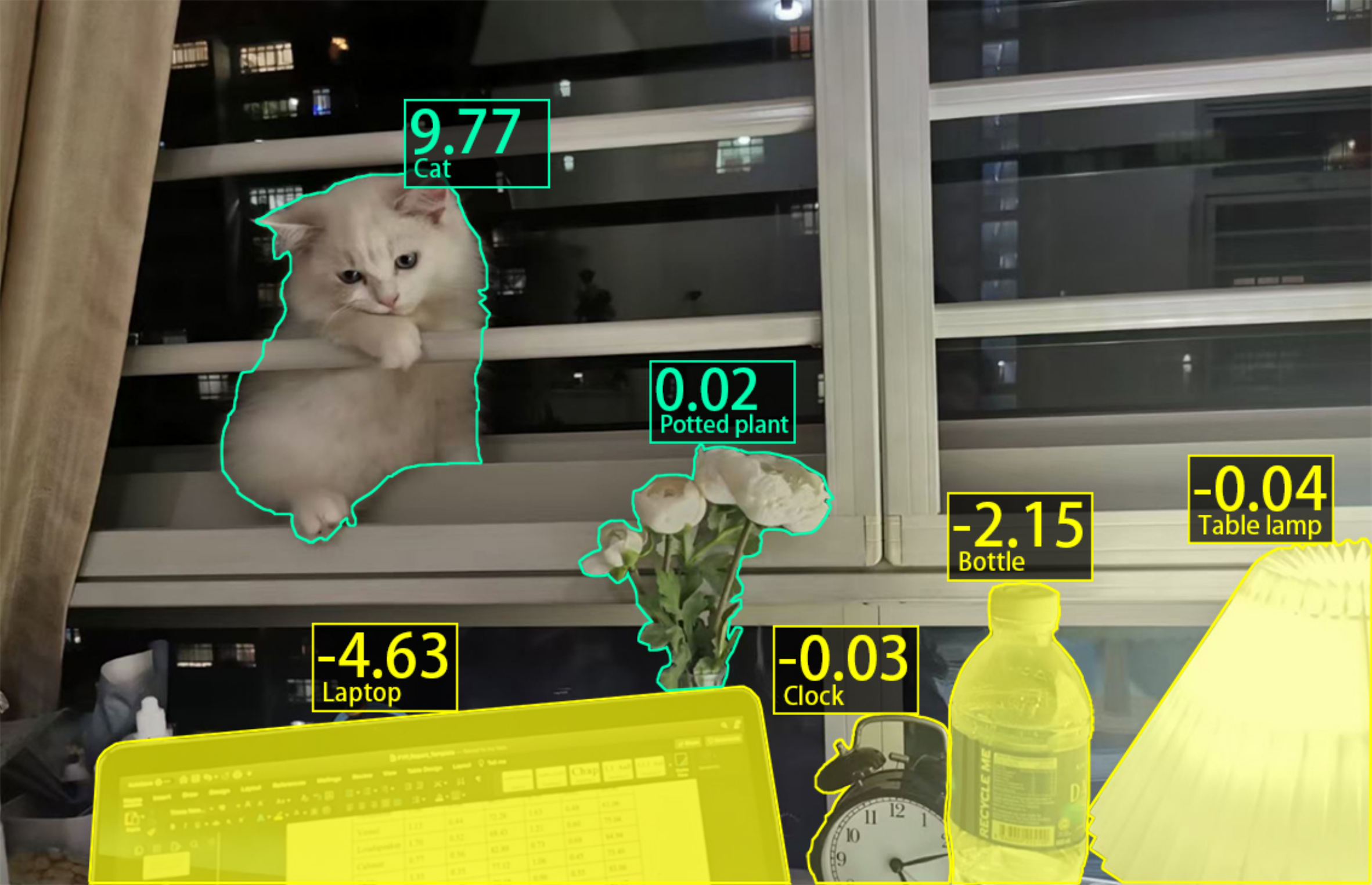}}
    \caption{Two examples of the performance of the contribution estimation algorithm. (a) $P18$ finds great results where the estimations are proportional to the size of the objects. (b) $P11$ likes most of the estimations but finds the estimation of an object (the bottle) does not align with her evaluation.}
    \label{fig:rank}
\end{figure}
We set the loss weight $\lambda_{\text{aes}}$ to $1$ to pay equal attention to aesthetics and content in the model. The model is trained for $100$ epochs with mini-batches of size $32$. Optimization is carried out by an Adam optimizer~\cite{kingma2014adam} with a learning rate of $4\times 10^{-4}$. An early stop mechanism is adopted to avoid model overfitting, and we stop training when the total loss does not decrease in $15$ consecutive epochs. We clip gradients with a norm larger than 5 to avoid dramatic network changes. A pre-trained ResNet with 101 layers is used for feature extraction. We do not use the fully-connected layers of the pre-trained ResNet model because the downstream task is different. The extracted feature maps are of the size $14\times 14\times 2048$. The extracted image features are also fed into the mixing network for generating the weights of the linear transformations. The mixing network is a two-layer fully-connected network with ReLU activation. The hidden layer has $128$ neurons.

The image inpainting model is trained on the ImageNet~\cite{deng2009imagenet} and Places2~\cite{zhou2017places} dataset. Corrupted images are the input to the generator, and the original images in the dataset are used as groundtruth for calculating the reconstruction loss and the discriminator loss. Similar to~\citet{zeng2020high}, we corrupt the images with two kinds of masks. The first is object masks, where we remove objects detected by Mask R-CNN from images. The second is random strokes to avoid a bias towards missing regions in the shape of objects. The generator consists of an encoder and a decoder. The encoder is a 6-layer convolutional neural network (CNN) with $\{48,48,96,96,192,192\}$ kernels at each layer. The decoder is also a CNN but has 7 convolutional layers with $\{192,192,96,96,48,$ $24,3\}$ kernels. An Adam optimizer with a learning rate of $1\times 10^{-4}$ is used to train mini-batches that contain 64 training samples.

\subsubsection{Setups of the User Studies for Modular Tests} To know the reaction of users to our clutter detection and removal algorithms, we design two user studies. Before the studies, we helped the participants install and get familiar with our app. The participants were encouraged to share their thoughts and usage experience with us anytime during the experiments. 

In the first user study, we focus on the clutter detection part, including the contribution of each object to the overall quality of the photo and the classification of clutter and normal objects. Experimenters are tasked to frame a scene and check the classification and contribution value of each object. They are encouraged to see the scene without masks by long-pressing the screen before drawing a conclusion. After checking the objects, they are invited to (1) double-click the objects whose classification seems to be improper; (2) point out cluttered objects whose contribution value does not align with their evaluation; and (3) answer a Likert question on a 7-point scale about their impression about the accuracy of clutter detection. We also conduct an open-ended interview with participants talking about their thoughts about the algorithm performance and the design of the interface. 

For the second user study, users are tasked to click the "clean" button and evaluate the image inpainting algorithm. They answer three Likert questions on a 7-point scale: (1) whether the generated content is visually plausible? (2) whether the generated content is semantically reasonable? (3) whether do you use the preview function frequently? We again conduct an open-ended interview in this study, exchanging opinions about the performance of the algorithm with the experimenters.

\subsubsection{Setups of the User Study for the System as a Whole} In this study, participants are asked to stage a scene using the default camera app before staging the same scene using our system. They are asked to favorite a single photo from these two. We also invite photography experts to compare the photos: for each pair of pictures, we present them to experts in a random order and ask them to select the one of higher quality. Moreover, experimenters are asked to answer three Likert questions, also on a 7-point scale, about whether they like the interactions, the visualization, and the overall guidance provided by our system. Both the experimenters and the experts are invited to take an interview where they can freely talk about their experience of using our app and, if any, their suggestions for improving the system.

\subsection{User Study for Modular Test}

\subsubsection{Clutter Detection}
In this study, we recruited 18 participants (10 male, 8 female) aged 19 to 30 years old ($\mu$=26). We observe clear results that favor our clutter detection results. In all the 18 photos shot by the participants, we detected 176 objects, 67 of which were classified as clutter. Among the 109 normal objects, experimenters thought 8 ($7.34\%$) of them should be clutter. For the cluttered objects, the category for 6 ($8.96\%$) of them was flipped by the author. These flips lead to an overall agreement degree of $92.05\%$ between our clutter detection algorithm and the judgment of experimenters. From an individual perspective, 6 participants did not alter any category, and $P1$ made the most (3) modifications. 

$P15$ finds the results (Fig.~\ref{fig:detection} (a)) accurate and help her make a decision about whether some visual elements should be present in the photo: "\emph{The color of the bus is harmonious with the cherry blossoms. This is what I want to stress in this photo. I am happy to see that the system thinks so. When shooting, I was not sure whether the lamps along the road should be present in my photo. The successful evaluation of the bus prompts me to trust the system and keep these lamps in the picture.}"

$P4$ reports that the masks draw his attention to clutter which is ignored beforehand (Fig.~\ref{fig:detection} (b)): "\emph{The striking masks caught my eyes. I didn't realize the bicycles were in the picture. Besides, the rider entered the viewport suddenly, and I should have waited for him to leave.}" $P4$ then waits for the car and the rider to move out of the picture and takes the second photo, using the trick of zooming in to focus on the major object and get rid of the bicycles. The experience of $P4$ indicates that our system meets the expectation for dealing with the second kind of clutter discussed in Sec.~\ref{sec:design}.

However, a similar detection result conflicts with the intention of $P1$ (Fig.~\ref{fig:detection} (c)): "\emph{I want to shoot a street scene, but the system classifies the cars as clutter. I am a bit frustrated because I don't think this is a hard case. The cars seem good within the background.}" Comparing the scenes of $P4$ and $P1$, we can see that our system usually classifies vehicles occupying a small part of the picture as clutter. This rule works fine for some pictures but does not generally hold. This case also highlights the complexity of the clutter detection task.

For the contribution estimation of cluttered objects, experimenters are encouraged to focus on the relative value. We note that there is no general standard for contribution estimation, and different people may hold different opinions regarding the relative contribution of various objects. To avoid ambiguity, we suggested beforehand that the participants only point out objects significantly conflicting with their evaluations. For the 69 cluttered objects after users' modification, participants found 7 ($10.15\%$) of them have an improper contribution estimation, among which 3 have a higher estimation.

This small number of conflicts suggests that participants generally agree with the predictions made by our algorithm. $P18$ supports that our system provides reasonable estimations: "\emph{In my photo of the avenue, there are four vehicles. The bigger the vehicle is, the lower the contribution value is. I like the results. They are intuitive. The bigger vehicle is indeed more distracting.}" (Fig.~\ref{fig:rank} (a))

$P11$ is satisfactory with most of the estimations but changes the order of the bottle in her picture (Fig.~\ref{fig:rank} (b)): "\emph{I think the bottle is the most distracting, both aesthetically and with respect to content. The laptop is also a distraction, but not as significant as the bottle.}" Although she does not agree with the relative contribution of the laptop and the bottle, she finds the system helps her improve her photographic work: "\emph{Despite the order, I agree that the laptop and the bottle should be kept out of the picture.}"

When rating the overall accuracy of clutter detection, participants do find our system provides a good classification that aligns well with their aesthetics (Mdn = 6, IQR = 6-7). These results indicate that the learned model well predicts the aesthetic experience of the general audience.

\begin{figure}
    \centering
    \subfigure[Photo taken by $P9$.]{\includegraphics[width=\linewidth]{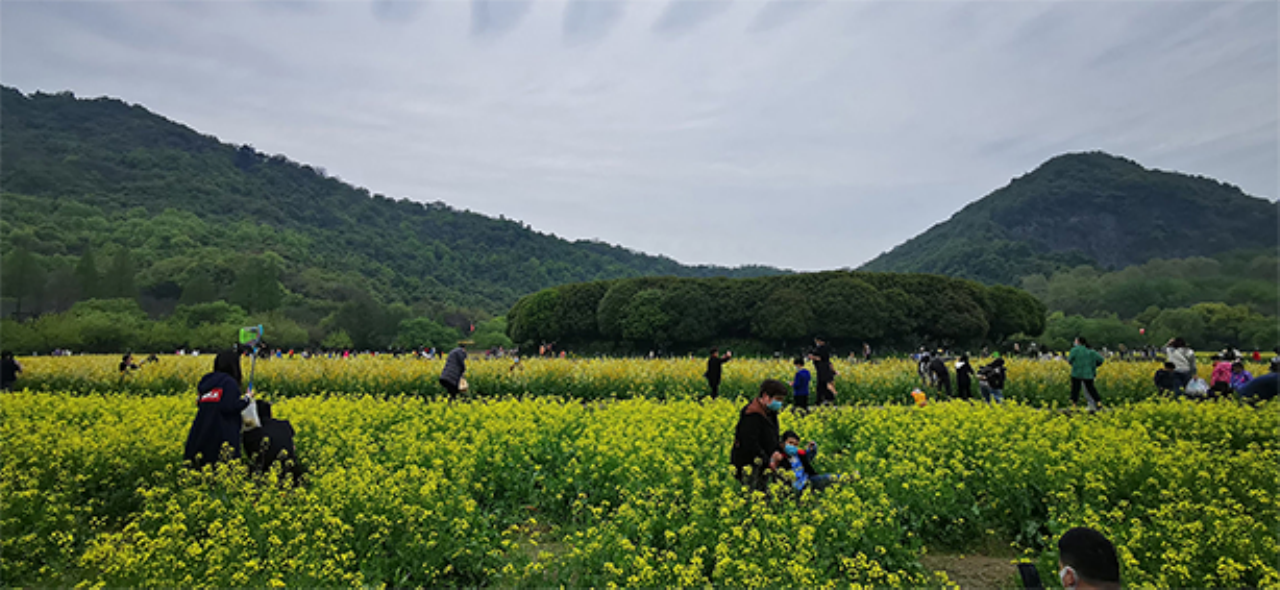}}
    \subfigure[$P9$'s photo after clutter removal.]{\includegraphics[width=\linewidth]{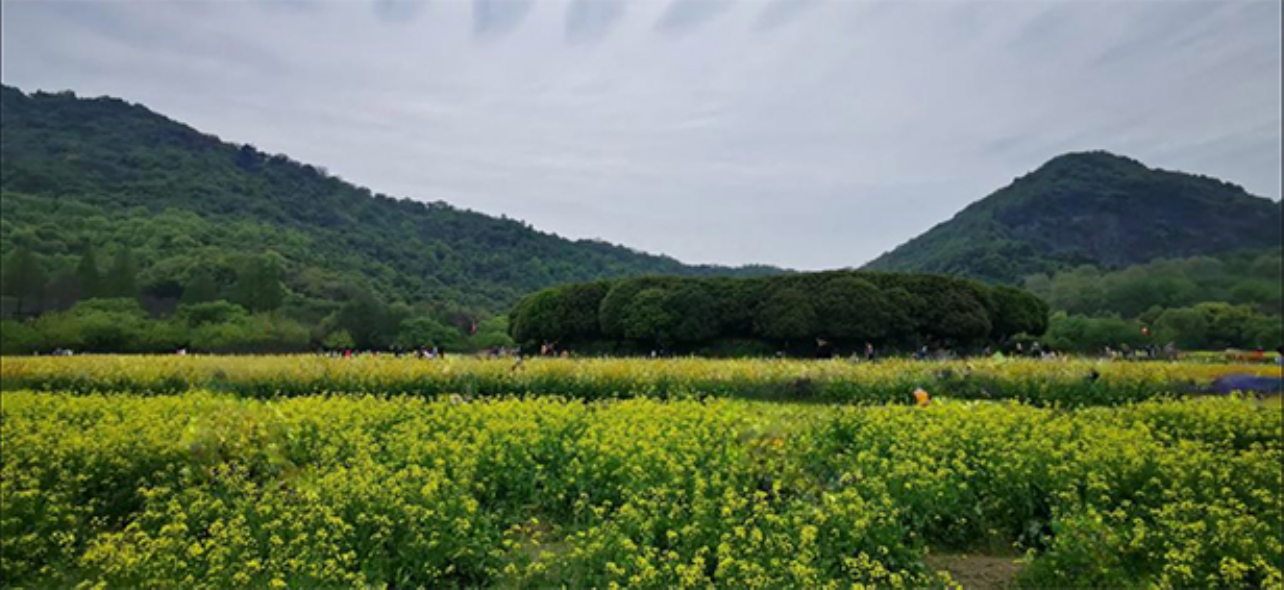}}
    \subfigure[Photo taken by $P2$.]{\includegraphics[width=0.49\linewidth]{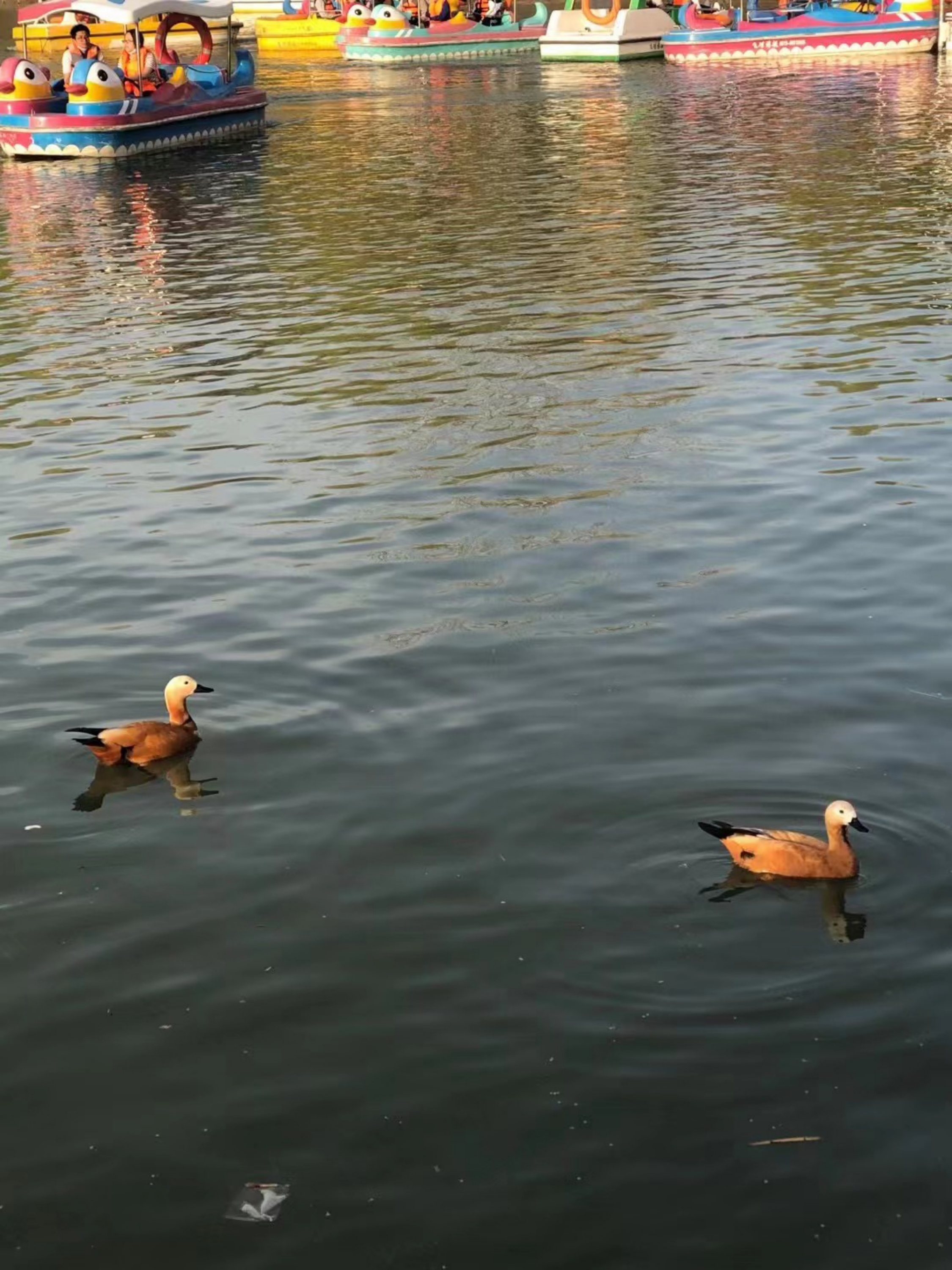}}\hfill
    \subfigure[$P2$'s photo after clutter removal.]{\includegraphics[width=0.49\linewidth]{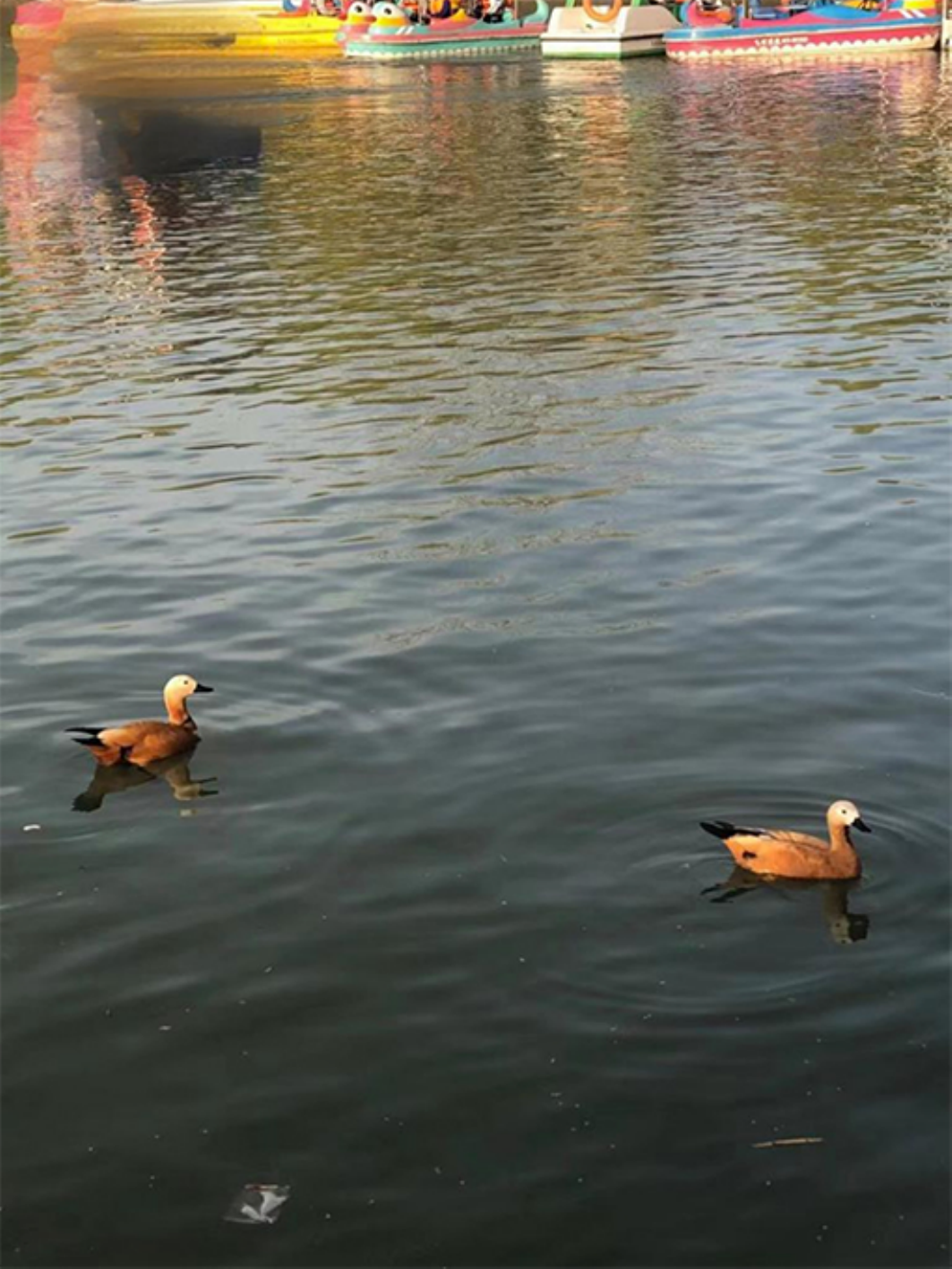}}
    \caption{Two examples of our clutter removal algorithms. (a-b) $P9$ finds the algorithm gets a superior result where the missing regions originally occupied by tourists are filled by the plants' textures in high quality. (c-d) A less satisfactory result encountered by $P2$. Artifacts and semantically unreasonable contents are found in the inpainted areas.}
    \label{fig:removal}
\end{figure}

\subsubsection{Clutter Removal}

The answers to the Likert questions reveal that the users find the inpainted images generally contain visually plausible (Mdn = 5, IQR = 5-6) and semantically reasonable (Mdn = 6, IQR = 5-7) content. The frequency of using the clutter removal function varies with different users (Mdn = 4, IQR = 3-6).

Most participants are satisfied with our image inpainting algorithm: "\emph{I like the photo after removing other tourists. To be honest, the result is far beyond my expectation. Even when I check the photo carefully, I can hardly distinguish the generated contents from real images.}" ($P9$) In this case (Fig.~\ref{fig:removal} (a-b)), the missing regions of other tourists are filled with realistic textures of plants in the background. However, we also notice that our algorithm struggles to deal with some complex scenarios where the background is occluded by a large object. For example, in Fig.~\ref{fig:removal} (c-d), we show a less satisfactory result where the lake and other boats are occluded by an object, and the reflection of this object further increases the difficulty of image inpainting. $P2$ gives a 4 for the "visually plausible" and "semantic reasonable" Likert questions because: "\emph{The generated part contains blur and an enlarged reflection of the original object. Although it is even hard for me to guess what is behind the boat, I think the artifacts make the photo less attractive.}"

The image inpainting algorithm is triggered 2.1 times on average for the staging of one scene, which is not frequent because the user can remove some cluttered objects using other approaches: "\emph{I didn't realize that an outdoor unit of air conditioners clutters my photo. This can be easily removed. I can change the camera angle, so I didn't use the "clean" button.}" ($P17$) This is in line with our design thoughts that the image inpainting algorithm is only used when removing the clutter in other ways takes a lot of efforts, for example, when ($P2$): "\emph{I have to use the clutter removal function. It was a holiday and there were many tourists there. It seems impossible that we find a clean scene without clicking the 'clean' button.}"

\subsection{User Study for the System}
The participants of this study are the same as in the modular test. We invite three photography experts, two photography graduate students (aged 24 and 25, respectively) and a college photography teacher (aged 49), to help evaluate the results.

We observe a significant preference of photos captured under the guidance of our system (Mdn = 6, IQR = 5-6) to the photos captured by the default camera app (Mdn = 4, IQR = 3-5) [Wilcoxon signed-rank test V = 3.5, p=0.007]. The photography experts also find the photos shot by our system are of higher quality (for the three experts, 15, 14, and 17 photos that are thought better by them are shot using our system).

As for the reasons why the participants prefer our system, some users \textbf{notice the clutter that they ignored}. For example, $P4$ realized that the bicycles are unrelated to the photo's theme (Fig.~\ref{fig:detection} (b)) and excluded them after being reminded by our system. Similarly, $P11$ was accustomed to her environment and included the bottle in the scene, which lowered the quality of her photo (Fig.~\ref{fig:rank} (b)). Our system found this problem and reminded her of it. $P2$ also appreciates that our system drew his attention to the small clutter in the upper part of the picture (Fig.~\ref{fig:removal} (c)).

More importantly, participants think our system \textbf{provides intuitive guidance on framing and compositions}. For example, $P15$ was not sure whether the lamps should be present (Fig.~\ref{fig:detection} (a)). Our system convinced her that the lamps contribute positively to the overall quality of the photo. $P4$ also made up his mind to include some vehicles in his photo after consulting our contribution estimations of cars (Fig.~\ref{fig:detection} (b)). $P11$ decided to exclude the laptop but keep the flower in her picture, which is also in line with the indication of our contribution estimations.

\begin{figure}
    \centering
    \subfigure[$P6$'s photo captured using the default camera app. Images of a car and a pillar on the glass clutter the photo.]{\includegraphics[width=\linewidth]{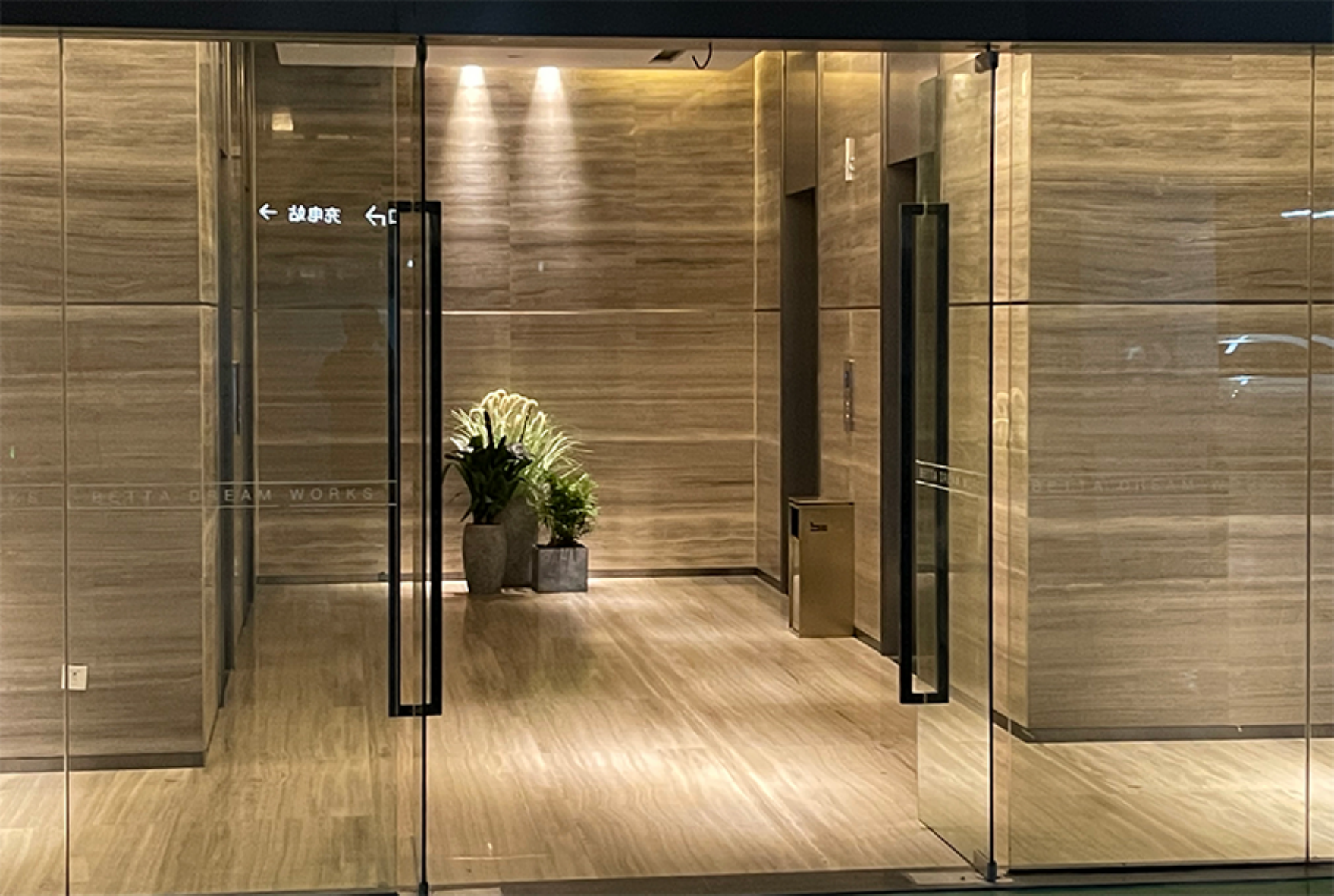}}
    \subfigure[Photo taken by $P6$ under the guidance of our system.]{\includegraphics[width=0.49\linewidth]{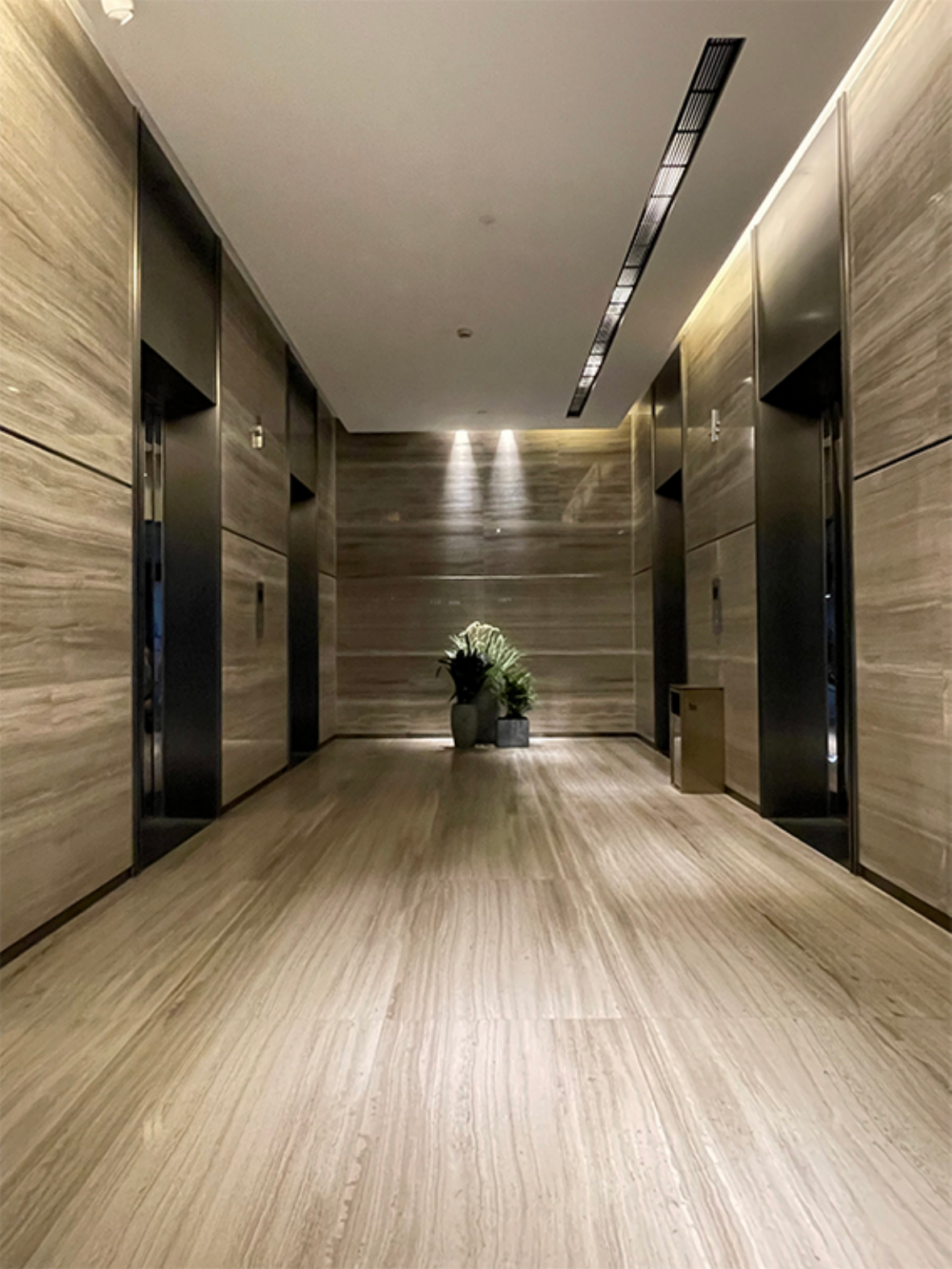}}\hfill
    \subfigure[$P6$'s photo after clutter removal.]{\includegraphics[width=0.49\linewidth]{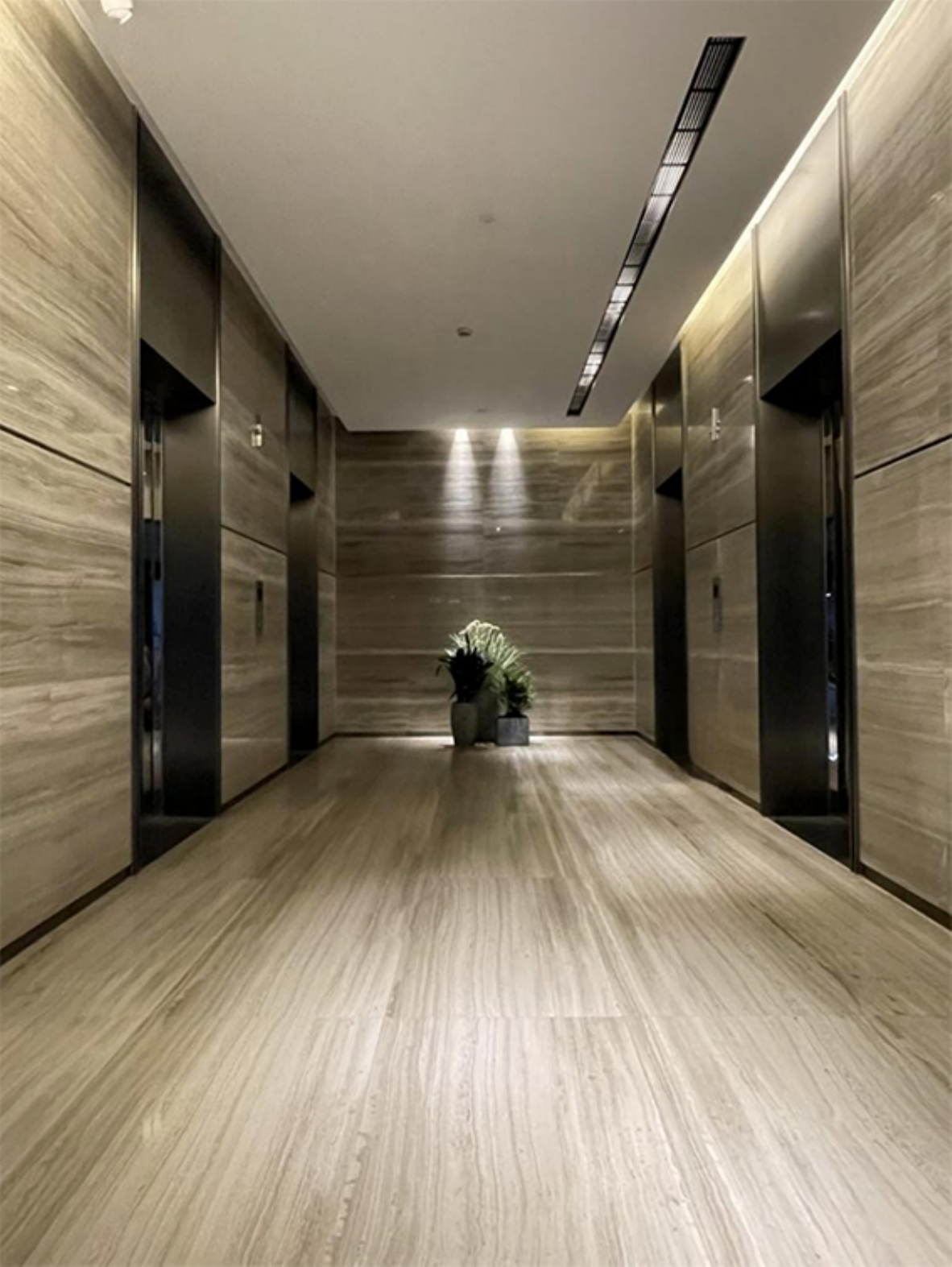}}
    \caption{Comparison of photos taken by $P6$ using the default camera app and our system. (a) Photos by the default camera. $P6$ is not aware that the images of a car and a pillar on the glass clutter the image. (b) Being reminded by our system, $P6$ excludes the glass in his photo. (c) He further finds the trash can causes imbalance. Since the trash can is immovable, he uses our image inpainting algorithm to get rid of it.}
    \label{fig:comparison}
\end{figure}

In addition, participants find \textbf{the image inpainting function gives opportunities to take photos in cluttered environments.} For example, $P9$ says: "\emph{Without the clean button, I may give up this photo} (Fig.~\ref{fig:removal} (a)). \emph{There was an endless stream of tourists, and I could hardly get a clean photo.}" $P6$ holds a similar opinion because, in his photo (Fig.~\ref{fig:comparison} (b)), the trash bin violates the balance of the photo but is immovable: "\emph{Normally, I would give up the scene. However, things are different in this system... That is amazing.}"

Some participants hold that \textbf{the system helps them save time and explore more.} $P9$ puts that: "\emph{With this app, I do not have to wait for other tourists. I spent my time there on shooting many other photos.}" $P15$ also thinks the system's contribution evaluation function is time-saving: "\emph{The system makes me confident about my photographic work. I used to spend a lot of time on photo compositions, but now I know trustful AI techniques are behind the system. I no longer need a long preparation time.}" 

Other participants find they \textbf{are motivated to be bolder in their photographic work}. For example, $P11$ says: "\emph{In my daily life, I like taking photos of dishes and desserts. I am always concerned that my photo contains much clutter when there are too many items on the table. I think the system does a good job in my cat photo} (Fig.~\ref{fig:rank} (b)), \emph{so I am more encouraged to try to capture photos of dishes. I believe the system can give good suggestions and guarantee that I do not make big mistakes.}"

\subsubsection{Interaction and Visualization} Most participants find our interactions user-friendly, attending to many aspects of clutter avoidance, and providing more opportunities to capture good photos. For example, $P6$ has a great experience using our system (Fig.~\ref{fig:comparison}): "\emph{I find the same picture appears differently in the tested system. The image of the car and the pillar on the glass are highlighted by masks. I didn't notice them in the default camera of the iPhone. To avoid these objects, I walked through the door to get a better scene. But the trash can destroyed the image balance here. Normally, I would give up the scene. However, things are different in this system, and I can remove the trash can simply by clicking a button. That is amazing.}"

$P7$ thinks that our system provides various functions using a few interactions: "\emph{When the system provider introduced how to interact, I thought the interactions are more than what I need. However, during shooting, every time I expect the system to have a function, it never let me down. I think the functions cover a large number of user needs with respect to clutter.}"

By contrast, participants give some suggestions for improving the visualization. $P1$ says: "\emph{I think the object masks are overly striking. I cannot see the original image clearly and have to press the screen for a long time.}" Another participant ($P4$) finds the pop-up box for suggestions is not aesthetically appealing. $P4$ suggests that "\emph{You may want to consider a pop-up box with a transparent background.}" These qualitative evaluations are also reflected in the results of the quantitative evaluation: most users are satisfied with our interactions (Mdn = 6, IQR = 6-7), and the rating for visualization is slightly lower (Mdn = 5, IQR = 5-6). Overall, the participants find the guidance provided by our system very helpful (Mdn = 6, IQR = 5-7).

        
        
        
        

%% file: 6-Conclusion.tex
\section{Conclusion}

In this paper, we introduce an in-camera capture-time guidance system that helps photographers identify and handle visual clutter in photos. This system is motivated by the observation that frequently appearing visual clutter often spoils the intended story of the photographer. We carry out a survey and classify clutter into three categories. We then develop a computational model to estimate the contribution of objects to the overall quality of photos, based on which we provide an interface that allows interactive identification of clutter. Suggestions and tools for dealing with different clutter are proposed. In particular, we develop an image inpainting algorithm to computationally cope with the clutter that is not cumbersome to remove. Users find the interactions provided by our system flexible, useful, and covering multiple aspects of clutter handling. For future work, we believe a generative model based on the current system that not only removes clutter but also proposes semantically reasonable and visually appealing image compositions is an exciting research direction.